\makeatletter\def\graphicscache@inhibit{true}\makeatother

\documentclass[10pt,twocolumn,letterpaper]{article}

 \usepackage{cvpr}              %
\usepackage[dvipsnames]{xcolor}

\definecolor{cvprblue}{rgb}{0.21,0.49,0.74}
\usepackage[pagebackref,breaklinks,colorlinks,citecolor=cvprblue]{hyperref}

\usepackage{graphicx}
\usepackage{amsmath}
\usepackage{amssymb}
\usepackage{booktabs}
\usepackage{threeparttablex}
\usepackage{tabu}
\usepackage{lipsum}
\usepackage{nccmath}
\usepackage{amsmath}
\usepackage{xfrac}

\usepackage{amsmath,scalerel}

\DeclareMathOperator*{\concat}{\scalerel*{\oplus}{\sum}}

\makeatletter
\@namedef{ver@everyshi.sty}{}
\makeatother

\usepackage{tikz}
\usetikzlibrary{positioning,3d,calc,intersections,spy,matrix,shapes}
\usetikzlibrary{quotes,arrows.meta}
\usetikzlibrary{positioning}
\usetikzlibrary{fit}

\makeatletter

\tikzdeclarecoordinatesystem{rel}{%
	\tikzset{cs/.cd,x=0pt,y=0pt,#1}%
	\pgfpointlineattime{(\tikz@cs@x / 100)}%
	{\pgfpointanchor{\tikz@pp@name{\tikz@cs@node}}{south west}}%
	{\pgfpointanchor{\tikz@pp@name{\tikz@cs@node}}{south east}}%
	\edef\tikz@cs@x{\the\pgf@x}%
	\pgfpointlineattime{(\tikz@cs@y / 100)}%
	{\pgfpointanchor{\tikz@pp@name{\tikz@cs@node}}{south west}}%
	{\pgfpointanchor{\tikz@pp@name{\tikz@cs@node}}{north west}}%
	\pgfpoint{\tikz@cs@x}{\pgf@y}%
}
\makeatother
\usepackage[export]{adjustbox}

\tikzset{font=\scriptsize\sffamily}

\pgfdeclarelayer{background}
\pgfdeclarelayer{foreground}
\pgfsetlayers{background,main,foreground}

\usepackage[pagebackref,breaklinks,colorlinks]{hyperref}

\usepackage[capitalize]{cleveref}
\crefname{section}{Sec.}{Secs.}
\Crefname{section}{Section}{Sections}
\Crefname{table}{Table}{Tables}
\crefname{table}{Tab.}{Tabs.}

\DeclareMathOperator{\Cov}{Cov}
\DeclareMathOperator{\Var}{Var}

\makeatletter
\renewcommand{\paragraph}{%
  \@startsection{paragraph}{4}%
  {\z@}{1.25ex \@plus 1ex \@minus .2ex}{-1em}%
  {\normalfont\normalsize\bfseries}%
}
\makeatother
\usepackage[accsupp]{axessibility} %

\def\paperID{6620} %
\def\confName{CVPR}
\def\confYear{2024}

\title{FSRT: Facial Scene Representation Transformer for Face Reenactment\\ from Factorized Appearance, Head-pose, and Facial Expression Features}

\author{Andre Rochow\\[-1pt]
	{\tt\small rochow@ais.uni-bonn.de}
	\and
	Max Schwarz\\[-1pt]
	{\tt\small schwarz@ais.uni-bonn.de}
	\and
	Sven Behnke\\[-1pt]
	{\tt\small behnke@cs.uni-bonn.de}
\end{tabular}\\[-3pt]
\begin{tabular}{c}
	{\small Autonomous Intelligent Systems - Computer Science Institute VI and Center for Robotics, University of Bonn, Germany}\\[-2pt]
	{\small Lamarr Institute for Machine Learning and Artificial Intelligence, Germany}
\end{tabular}\\[-\baselineskip]
\begin{tabular}{c}
}

\begin{document}
\maketitle

\begin{abstract}
The task of face reenactment is to transfer the head motion and facial expressions from a driving video to the appearance of a source image, which may be of a different person (cross-reenactment). Most existing methods are CNN-based and estimate optical flow from the source image to the current driving frame, which is then inpainted and refined to produce the output animation. We propose a transformer-based encoder for computing a set-latent representation of the source image(s). We then predict the output color of a query pixel using a transformer-based decoder, which is conditioned with keypoints and a facial expression vector extracted from the driving frame. Latent representations of the source person are learned in a self-supervised manner that factorize their appearance, head pose, and facial expressions. Thus, they are perfectly suited for cross-reenactment. In contrast to most related work, our method naturally extends to multiple source images and can thus adapt to person-specific facial dynamics. We also propose data augmentation and regularization schemes that are necessary to prevent overfitting and support generalizability of the learned representations. We evaluated our approach in a randomized user study. The results indicate superior performance compared to the state-of-the-art in terms of motion transfer quality and temporal consistency.\footnote{Code \& Video: \url{https://andrerochow.github.io/fsrt}}

\end{abstract}

\section{Introduction}\label{sec:intro}

Face reenactment is a special case of the motion transfer task~\citep{monkey_net,fom}.
Its objective is to synthesize a realistic facial animation combining the appearance given by one or more images of a source person and the facial expression and head motion of a driving video, which may be of a different person.
The driving frame is used to transform the source image to the desired expression and head pose.
When the driving video is of the same person (self-reenactment), applications include, e.g., low-bandwidth video conferencing.
The more interesting and challenging case is when the driving video is of a different person (cross-reenactment) since, if successful, only one or few images of the source person are required to create a realistic facial animation.
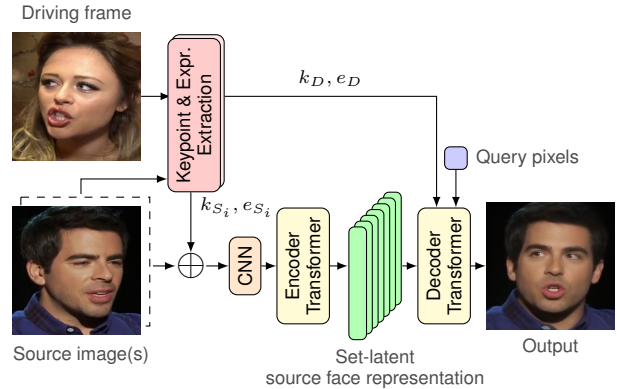
\begin{figure}
	\centering\newlength\teaserimgh\setlength\teaserimgh{1.7cm}
	\begin{tikzpicture}[
	every node/.style={align=center},
	every label/.style={inner sep=3pt,text=black!70},
	n/.style={draw=black,fill=yellow!20,rounded corners=3pt},
	img/.style={inner sep=0pt},
	]
	\node[img] (img1) at (0,0,0) {\includegraphics{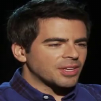}};
	
	\begin{pgfonlayer}{background}
	\coordinate (dashnortheast) at ($(img1.north east)+(0,0,-0.3)$);
	\draw[dashed] ($(img1.north west)+(0,0,-0.3)$) rectangle ($(img1.south east)+(0,0,-0.3)$);
	\end{pgfonlayer}

	\node[inner sep=1pt, fit=(img1)(dashnortheast),label={below:Source image(s)}] (imgbound) {};
	
	\node[img,anchor=south west,label={above:Driving frame}] (drv) at ($(img1.north west)+(0.0,0.6)$) {\includegraphics{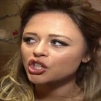}};
	\node[n,rotate=90,anchor=north,fill=red!10] (kp_) at ($(drv.east)+(0.4,-0.2)$) {Keypoint \& Expr.\\Extraction};
	\node[n,rotate=90,anchor=north,fill=red!20] (kp) at ($(drv.east)+(0.35,-0.25)$) {Keypoint \& Expr.\\Extraction};

	\node[right=0.3cm of imgbound,font=\Large,inner sep=0pt] (concat) {$\oplus$};
	
	\node[right=0.3cm of concat,n,fill=orange!20] (cnn) {\adjustbox{rotate=90}{CNN}};
	
	\node[right=0.2cm of cnn,rotate=90,anchor=north,n] (enc) {Encoder\\Transformer};
	
	\coordinate (c) at ($(enc.south)+(0.6,0)$);
	\begin{scope}[
	shift={(c)},
	every node/.append style={
		n,fill=green!30,minimum height=1.5cm,minimum width=0.2cm,rounded corners=2pt
	}]
	
	\def\fitlist{}
	\foreach \z in {-3,...,3} {
		\pgfmathparse{0.2 * \z}
		\node (r\z) at (0,0,\pgfmathresult) {};
		\xdef\fitlist{\fitlist(r\z)}
	}
	\end{scope}
	\node[inner sep=0pt, fit=\fitlist,label={below:Set-latent\\source face representation}] (latents) {};
	
	\node[n,right=0.2cm of latents,anchor=north,rotate=90] (dec) {Decoder\\Transformer};
	
	\node[anchor=south west,n,minimum width=0.3cm,minimum height=0.3cm,rounded corners=2pt,fill=blue!20,label={right:Query pixels}] (qpix) at ($(dec.east)+(0,0.5)$) {};
	
	\node[img,right=0.2cm of dec.south,label={below:Output}] (out) {\includegraphics{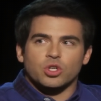}};

	\begin{pgfonlayer}{background}
	\draw[-latex] (drv) -- (drv-|kp_.north);
	\end{pgfonlayer}
	\draw[-latex] (drv-|kp_.south) -| ($(dec.east)+(-0.1,0)$) node [pos=0.25,above] {$k_D, e_D$};
	
	\coordinate (c2) at ($(imgbound.north)+(0,0.2)$);
	\draw[-latex] (imgbound.north) -- (c2) -- ($(c2-|kp.north)+(0,0)$);
	\draw[-latex] (concat|-kp.west) -- (concat) node [pos=0.25,right] {$k_{S_i},e_{S_i}$};
	
	\draw[-latex] (imgbound) -- (concat);
	\draw[-latex] (concat) -- (cnn);
	\draw[-latex] (cnn) -- (enc);
	\draw[-latex] (enc) -- (latents);
	\draw[-latex] (latents) -- (dec);
	\draw[-latex] (dec) -- (out);
	\draw[-latex] (qpix) -- (qpix|-dec.east);
	\end{tikzpicture} \vspace{-1ex}
	\caption{Overview of our method (relative motion transfer). The source image(s) are encoded along with keypoints $k_S$, capturing head pose, and facial expression vectors $e_S$ to a set-latent representation of the source person. The decoder attends this  representation for a query pixel, conditioned on keypoints $k_D$ and a facial expression vector $e_D$ extracted from the driving frame. $\oplus$ denotes pixel-wise concatenation. Images from the VoxCeleb test set~\cite{voxceleb}.
	}
	\label{fig:teaser} \vspace{-1ex}
\end{figure}

Most face reenactment methods are CNN-based~\citep{monkey_net,fom,dagan,spline,one-shot,dagan,hsu2022dual,fewshotvid2vid,zhao2021sparse, zakharov2019few,ha2020marionette,wiles2018x2face}; and many of these utilize optical flow between the source and driving frames for morphing the source image, followed by a refinement stage~\citep{monkey_net,fom,dagan,spline,one-shot}.

Inspired by recent successes in scene reconstruction~\citep{srt,osrt}, we apply a transformer-based architecture to face reenactment that encodes the face of the source person as a set of latent vectors (see \cref{fig:teaser}). This representation is learned in a self-supervised way.
We then sample each target pixel location with a transformer-based decoder conditioned on keypoints and an expression vector that are extracted from the driving frame.
The learned set-latent representation 
of the source person factorizes their head pose, appearance, and facial expression, which enables accurate head motion and facial expression transfer, also for cross-reenactment.

While our method yields state-of-the-art results in absolute motion transfer (i.e., when the driving keypoints are used unmodified),
it especially increases robustness in the case of relative motion transfer~\citep{fom}, a mode which reduces unwanted
leaking of face shape from the driving frame.
Relative motion transfer is initialized by finding a source and driving frame with well-matched head poses and
expressions and then operates by applying motions to the source frame in relative fashion.

Many methods encode facial expression by keypoints,
which are augmented with local spatial transformations~\citep{fom,dagan,spline}. Here, relative motion is encoded as relative transforms from the initial driving frame and is applied
in the source frame.
Of course, this is highly sensitive to the initialization. %
By decoupling head pose from expression and describing expression in an absolute manner, our approach reduces the influence of initialization on expression transfer.

Finally, we note that our approach makes few assumptions since there is no explicit model of motion
or correspondence. Instead, the set-latent representation of the source person is learned in a self-supervised way using conditioning with keypoints and a facial expression vector from the source on the encoder and from the driving frame on the decoder, respectively.

\noindent In summary, our contributions include:
\begin{enumerate}
	\item A novel transformer-based architecture for face reenactment which learns a
	global set-latent representation of source images and allows rendering conditioned on keypoints and expression vectors,
    \item a latent expression description invariant to appearance, head shape, and head pose, greatly improving cross-reenactment,
	\item augmentation and regularization methods for training that support the separation of facial expression information (expression vector), appearance (set-latent representation), and head pose (keypoints) without overfitting,
	\item application of adversarial and perceptual losses to scene representation transformers, which greatly improves realism and sharpness,
	\item a detailed evaluation, where we outperform state-of-the-art in motion transfer quality and temporal consistency, including a user study which also shows superiority in subjective preference trials.
\end{enumerate}

\section{Related Work}

\paragraph{Face Image Synthesis.}

Face image synthesis deals with the generation of new non-existing faces~\cite{hou2019improving} 
even from text input~\cite{lin2023catch,huang2023collaborative}, completion of missing facial regions of known faces~\cite{li2017generative,zhou2020learning}, or manipulation of expression and appearance of known faces in manual~\cite{jo2019sc,CelebAMask,huang2023collaborative} or automatic manner~\cite{ganimation,stargan,deng2020disentangled,shen2020interpreting}.

In contrast to these methods, our approach combines the continuously changing head pose and expression from the driving video with the appearance of the source, so that a natural and temporally consistent video stream is produced.

\paragraph{Talking Head Synthesis and Face Reenactment.}
Talking Head Synthesis aims to make head poses, emotions, and especially speech controllable. Here, lip movement is mainly reconstructed from audio~\cite{wang2021audio2head,zhou2021pose,zhou2019talking,wiles2018x2face,richard2021meshtalk,wang2023progressive,wang2023lipformer,wang2023seeing,ma2023styletalk,xu2023high}. Closely related, 
face reenactment~\cite{thies2016face2face}, which is a motion transfer task, aims to apply the motion given by a driving frame to the appearance defined by a source image.
An even more challenging problem is VR facial animation, where the driving face is additionally occluded by a head-mounted display (HMD)~\cite{lombardi2018deep,facevr,vr_facial,faceaudio,rochow2022vr,rochow2023attention}. Here, especially the alignment problem between facial images and mouth camera images makes it difficult to generate training data. 

Generally, there are different types of driving information being used. Where some methods only utilize facial keypoints~\cite{monkey_net,hsu2022dual,fewshotvid2vid,zhao2021sparse, zakharov2019few}, 
other methods additionally use image features from a driving frame~\cite{fom,dagan,spline,one-shot,siarohin2021motion,ha2020marionette}. Also, audio can be used if available~\cite{agarwal2023audio}.

Some methods~\cite{hsu2022dual,zhao2021sparse} utilize external 3D keypoints extraction~\cite{bulat} for face reenactment. \citet{hsu2022dual} use a separate generator to predict more accurate driving keypoints, given initial keypoints and a source image. \citet{monkey_net} learn keypoints self-supervised and use them to predict a deformation grid of the source image into the driving keypoints. To resolve keypoint ambiguities, \citet{fom} estimate local affine transformations into a canonical space for each keypoint area. This allows far more facial expressions to be represented. Based on~\cite{fom}, \citet{dagan} learn depth maps, which they use to predict more accurate keypoints and
dense depth-aware attention maps, which can attend to important semantic facial areas. \citet{spline} use a motion estimation based on thin-plate spline transformations to produce a more flexible optical flow. They use multi-resolution depth maps and occlusion masks to inpaint missing regions more realistically. %

However, driving motion does not necessarily has to be described by keypoints \cite{siarohin2021motion,thies2016face2face,wiles2018x2face,dpe}. \citet{wiles2018x2face} directly predict the sampling coordinates to a canonical embedding of a face. A separate driving network then predicts the mapping from the embedded face to the driving frame. \citet{siarohin2021motion} bypass the keypoint estimation step from predicted heatmaps and consider them as regions. They estimate the principal components of the region to predict an in-plane rotation and scaling, which is used to estimate a more accurate pixel-wise optical flow. \citet{dpe} learn a disentanglement of pose and expression, so that different driving sources can be used.  Very recently, \citet{li2023one} learned to embed a source image into a canonical volume and predict the deformation of individual sampled rays to estimate the optical flow.

\citet{wang2022latent} and \citet{gong2023toontalker} replace the motion network proposed by \citet{monkey_net} with custom modules (Linear Motion Decomposition and a transformer module enabling domain switching, respectively), but remain fundamentally based on CNNs (in encoder and decoder) and a warp-and-refine architecture.

Unlike related methods, we use a transformer-style architecture to predict a latent scene representation of the source images and learn expression vectors which are decoupled from appearance and head pose information. Instead of modeling optical flow and motion explicitly, we learn to attend the latent scene representation, with keypoints and latent expression vectors extracted from a driving frame.
This allows us to generate results of higher accuracy, while significantly improving the temporal consistency.

\paragraph{Scene Representation Transformers.}
While transformers \cite{vaswani2017attention} were originally developed for natural language processing, vision transformers have also been highly successful \citep{dosovitskiy2021image,liu2021swin}. %
Recently, \citet{srt} proposed Scene Representation Transformers (SRT) to learn an internal scene representation encoded in a set of latents vectors. Given these latent vectors and a camera pose, SRT allow novel-view rendering without explicitly modeling the scene geometry. Based on this, \citet{osrt} propose a slot attention module to instead predict an object-centric slot scene representation, in which different objects are separated without any supervision.

In this work, we reformulate SRT~\cite{srt} for the face reenactment task and demonstrate how to efficiently model and query dynamics in the learned face representation. Unlike \cite{srt}, we aim to reconstruct photorealistic faces. To this end, we propose training with perceptual~\cite{perceptual} and adversarial losses~\cite{gan}, which significantly improves the output quality.

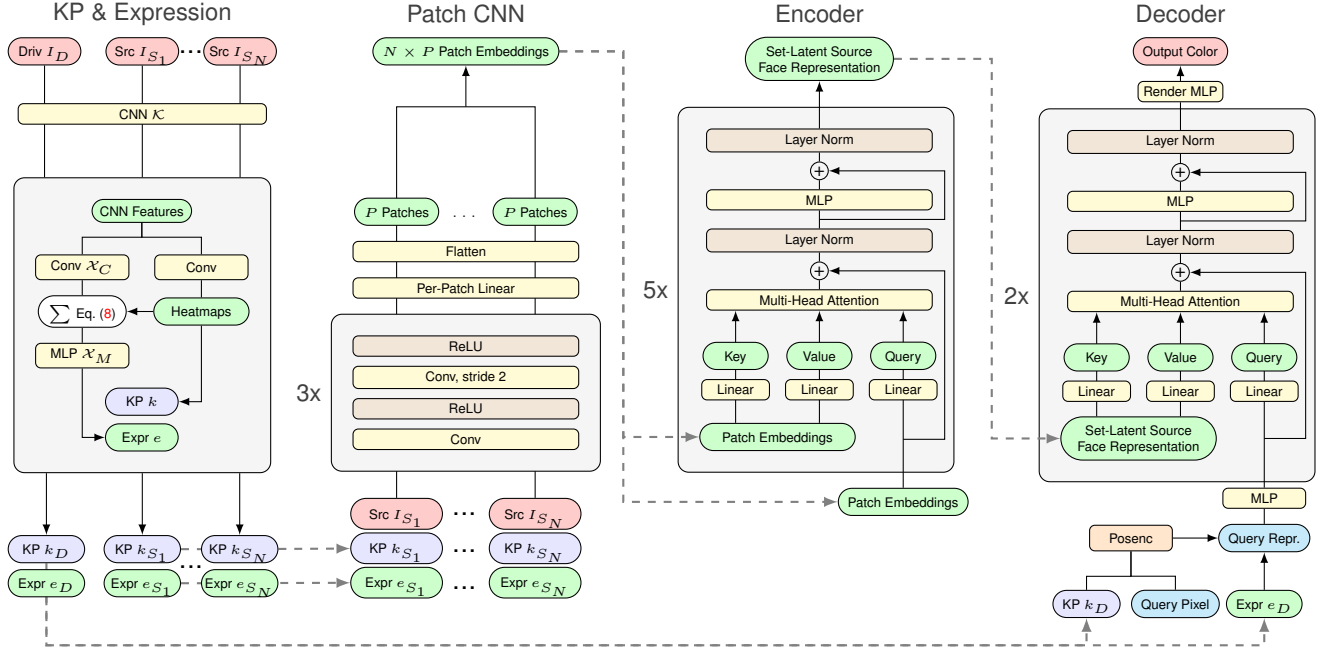
\begin{figure*}[htb!]
	\centering
	\begin{tikzpicture}[
	font=\sffamily\tiny,
	basic/.style={draw=black},
	data/.style={basic,rounded rectangle,text depth=0pt,fill=green!20},
	imgdata/.style={data,fill=red!20},
	kpdata/.style={data,fill=blue!10},
	query/.style={data,fill=cyan!20},
	module/.style={basic,rounded corners=2pt,inner sep=2pt},
	lmodule/.style={module,fill=yellow!20},
	encmodule/.style={basic,rounded corners=2pt, fill=orange!20,inner sep=3pt},
	mlp/.style={basic,rounded corners=2pt, fill=yellow!20,inner sep=2pt},
	smodule/.style={module,fill=brown!20},
	repeatbox/.style={
		inner sep=8pt,draw=black,rounded corners,fill=black!4,
		prefix after command={\pgfextra{\tikzset{every label/.append style={font=\sffamily\small,text=darkgray}}}},
	},
	plus/.style={circle,draw=black,inner sep=0.5pt,font=\scriptsize}
	]
	\begin{scope}[shift={(-4,0)}]
	\matrix[matrix of nodes,anchor=south,row sep=0.2cm] (x) {
		|[lmodule,minimum width=1.24cm] (convx)| Conv $\mathcal{X}_C$ \\
		|[data, fill=white] (sum)| \scriptsize $\sum$ \tiny  \cref{eq:8} \\
		|[lmodule,,minimum width=1.24cm] (mlpx)| MLP $\mathcal{X}_M$ \\
	};
	\node[lmodule, right=0.33cm of convx,minimum width=1.24cm, inner sep =2.8pt] (heatconv) {Conv};
	\node[data,minimum width=1.5cm] at ($(sum-|heatconv)$) (heat) {Heatmaps};
	\node[kpdata,minimum width=1.2cm] at ($(sum)!0.5!(heat)+(0,-1.2)$) (kpinside) {KP $k$};
	\node[data,minimum width=1.2cm, below=0.1 of kpinside] (exprinside) {Expr $e$};
	\node[minimum width=0pt, above=0.15 of convx] (dummy_convx) {};

	\node[data, minimum width=1cm,inner sep = 2.5pt] at ($(convx)!0.5!(heatconv)+(0,0.75)$) (featinblock) {CNN Features};

	\begin{pgfonlayer}{background}
	\node[repeatbox,fit=(mlpx)(heat)(convx)(exprinside)(dummy_convx)(featinblock),label={}] (expblock) {}; %
	\end{pgfonlayer}

	\node[kpdata,below=0.823cm of expblock,minimum width=1.25cm] (kpssrc1) {KP $k_{S_1}$};
	\node[kpdata,right= 0.265cm of kpssrc1,minimum width=1.25cm] (kpssrc) {KP $k_{S_N}$};
	\node[kpdata,left= 0.245cm of kpssrc1,minimum width=1.25cm] (kpsdriv) {KP $k_{D}$};
	\node[minimum width=0pt] at ($(kpssrc1)!0.5!(kpssrc)+(0.02,-0.23)$) (dots_out) {\scriptsize$\textbf{\dots}$};
	\node[data,below=0.078cm of kpssrc,minimum width=1.25cm] (latentsinput) {Expr $e_{S_N}$};
	\node[data,below=0.078cm of kpssrc1,minimum width=1.25cm] (latentsinput1) {Expr $e_{S_1}$};
	\node[data,below=0.078cm of kpsdriv,minimum width=1.25cm] (latentsdiving) {Expr $e_{D}$};

	\node[imgdata,minimum width=1.2cm] at ($(kpssrc1)+(0,6.623)$) (xinputimages1) {Src $I_{S_1}$};
	\node[imgdata, right= 0.32cm of xinputimages1,minimum width=1.2cm]  (xinputimages) {Src $I_{S_N}$};
	\node[imgdata, left= 0.32cm of xinputimages1,minimum width=1.2cm] (xdrivingimage) {Driv $I_D$};
	\node[minimum width=0pt] at ($(xinputimages)!0.5!(xinputimages1) + (0.015,0)$) (dots_inp) {\scriptsize$\textbf{\dots}$};
	\node[lmodule, minimum width=3.3cm, below=0.5cm of xinputimages1] (cnnk) {CNN $\mathcal{K}$};

	\node[text=darkgray] at ($(xinputimages)!0.5!(xdrivingimage)+(0,0.5)$) (text) {\sffamily\small KP \& Expression};

	\node[] (dummy_convheatconv) at ($(convx)!0.5!(heatconv)+(0,0.75)$) {};
	
	\begin{pgfonlayer}{background}
	\draw[-] (xinputimages) --(xinputimages|-expblock.north);
	\draw[-] (xdrivingimage) --(xdrivingimage|-expblock.north);
	\draw[-latex] (kpssrc1|-expblock.south) -- (kpssrc1);
	\draw[-latex] (kpssrc|-expblock.south) -- (kpssrc);
	\draw[-latex] (kpsdriv|-expblock.south) -- (kpsdriv);

	\draw[-] (xinputimages1) --(xinputimages1|-expblock.north);
	\draw[-] (dummy_convheatconv) --++ (0,-0.35) -|(convx);
	\draw[-] (dummy_convheatconv) --++ (0,-0.35) -|(heatconv);
	\draw[-] (heatconv) -- (heat);
	\draw[-latex] (heat) |- (kpinside);
	\draw[-latex] (heat) -- (sum);
	\draw[-] (convx) -- (sum);
	\draw[-] (sum) -- (mlpx);
	\draw[-latex] (mlpx) |- (exprinside);
	\end{pgfonlayer}
	\end{scope}

	\begin{scope}[shift={(1.1,-1.25)}, module/.append style={minimum width=3cm}]
	\node[matrix, matrix of nodes, row sep=0.12cm, anchor=south,inner sep=8pt,repeatbox,label={left:3x}] (cnnblock) {
		|[smodule] (cnnrelu2)| ReLU \\
		|[lmodule] (cnnconv2)| Conv, stride 2 \\
		|[smodule] (cnnrelu1)| ReLU \\
		|[lmodule] (cnnconv1)| Conv \\
	};
	
	\node[below=0.2cm of cnnblock,matrix, nodes={minimum width=1.5cm}, inner sep= 4pt] (cnninputs) {
		\node[imgdata] (cnninputimg1) {Src $I_{S_1}$};
		\node[kpdata,below=1pt of cnninputimg1] (cnninputkp1) {KP $k_{S_1}$};
		\node[data,below=1pt of cnninputkp1] (cnninputexpr1) {Expr $e_{S_1}$};
		&
		\node[minimum width=0pt] at (0,-0.05) (dots1) {\scriptsize$\textbf{\dots}$};
		\node[minimum width=0pt,below=4.5pt of dots1] (dots2) {\scriptsize$\textbf{\dots}$};
		\node[minimum width=0pt,below=4.5pt of dots2] {\scriptsize$\textbf{\dots}$};
		&
		\node[imgdata] (cnninputimg2) {Src $I_{S_N}$};
		\node[kpdata,below=1pt of cnninputimg2] (cnninputkpN) {KP $k_{S_N}$};
		\node[data,below=1pt of cnninputkpN] (cnninputexprN) {Expr $e_{S_N}$};
		\\
	};
	
	\node[above=0.2cm of cnnblock, lmodule] (cnnplin) {Per-Patch Linear};
	\node[above=0.2cm of cnnplin, lmodule] (cnnflat) {Flatten};
	
	\coordinate (flatstart) at ($(cnnflat.north)+(0,0.2)$);
	
	\node[data,anchor=south] (cnnpatch1) at (flatstart-|cnninputimg1) {$P$ Patches};
	\node[anchor=south] at (flatstart) {$\dots$};
	\node[data,anchor=south] (cnnpatch2) at (flatstart-|cnninputimg2) {$P$ Patches};
	\node[inner sep=0pt,fit=(cnnpatch1)(cnnpatch2)] (cnnpatch) {};
	
	\node[data] at ($(xdrivingimage-|cnnpatch)$) (cnnpatchemb) {$N \times P$ Patch Embeddings};
	
	\node[text=darkgray] at ($(cnnpatchemb|-text)$) (textcnn) {\sffamily\small Patch CNN};

	\begin{pgfonlayer}{background}
	\coordinate (join) at ($(cnnpatchemb.south)+(0,-0.55)$);
	\draw (cnninputimg1) |- (join);
	\draw (cnninputimg2) |- (join);
	\draw[-latex] (join) -- (cnnpatchemb);
	\end{pgfonlayer}
	\end{scope}

	\begin{scope}[shift={(5.8,-1.18)},module/.append style={minimum width=3cm}]
	\matrix[
	every outer matrix/.style={inner sep=8pt},anchor=south, matrix of nodes, row sep=0.12cm, column sep=0.2cm,
	module/.append style={minimum width=0.9cm},
	data/.append style={minimum width=1.1cm},
	] (encatt) {
		|[data]| Key       & |[data]| Value     & |[data]| Query \\
		|[lmodule](enclin1)| Linear & |[lmodule](enclin2)| Linear & |[lmodule](enclin3)| Linear \\[0.2cm]
		|(encp1)[module,draw=none,minimum width=0cm]| \phantom{Linear} & |(encp2)[module,draw=none,minimum width=0cm]| & \coordinate (encp3); \\
	};
	\node[data,minimum width=2.3cm] (encpe) at ($(encp1)!0.5!(encp2)$) {Patch Embeddings};
	
	\matrix[above=0cm of encatt,matrix of nodes, row sep=0.12cm] {
		\node[smodule] (encnorm2) {Layer Norm}; \\
		\node[plus] (encplus2) {+}; \\
		\node[lmodule](encmlp) {MLP}; \\
		\coordinate (encres); \\
		\node[smodule](encnorm1) {Layer Norm}; \\
		\node[plus] (encplus1) {+}; \\
		\node[lmodule](encmha) {Multi-Head Attention}; \\
	};
	
	\begin{pgfonlayer}{background}
	\node[repeatbox,fit=(encnorm2)(encpe),label={left:5x}] (encblock) {};
	\end{pgfonlayer}
	
	\node[data,anchor=north] (encpe2) at ($(encblock.south-|enclin3)+(0,-0.2)$) {Patch Embeddings};
	
	\node[data,align=center] (encscene) at ($(xdrivingimage-|encnorm2)+(0,-0.1)$) {Set-Latent Source\\Face Representation};
	
	\node[text=darkgray] at ($(encscene|-text)$) (textenc) {\sffamily\small Encoder};
	
	\begin{pgfonlayer}{background}
	\draw[-latex] (enclin1|-encpe.north) -- (enclin1|-encmha.south);
	\draw[-latex] (enclin2|-encpe.north) -- (enclin2|-encmha.south);
	\draw[-latex] (encpe2) -- (encpe2|-encmha.south);
	\draw[-latex] (encmha) -- (encplus1) -- (encnorm1) -- (encmlp) -- (encplus2) -- (encnorm2) -- (encscene);
	
	\coordinate (encr) at ($(encblock.east)+(-0.13,0)$);
	\draw[-latex] (encp3) -- (encp3-|encr) |- (encplus1);
	\draw[-latex] (encres) -- (encres-|encr) |- (encplus2);
	\end{pgfonlayer}
	\end{scope}
	
	\begin{scope}[shift={(10.6,-1.2)},module/.append style={minimum width=3cm}]
	\matrix[
	every outer matrix/.style={inner sep=8pt},anchor=south, matrix of nodes, row sep=0.12cm, column sep=0.2cm,
	module/.append style={minimum width=0.9cm},
	data/.append style={minimum width=1.1cm},
	] (decatt) {
		|[data]| Key       & |[data]| Value     & |[data]| Query \\
		|[lmodule](declin1)| Linear & |[lmodule](declin2)| Linear & |[lmodule](declin3)| Linear \\[0.2cm]
		|(decp1)[module,draw=none,minimum width=0cm]| \phantom{Linear} & |(decp2)[module,draw=none,minimum width=0cm]| & |[inner sep=0pt](decp3)| \\
	};
	\node[data,minimum width=2.3cm,align=center] (decscene) at ($(decp1)!0.5!(decp2)$) {Set-Latent Source\\Face Representation};
	
	\matrix[above=0cm of decatt,matrix of nodes, row sep=0.12cm] (decblock) {
		\node[smodule] (decnorm2) {Layer Norm}; \\
		\node[plus] (decplus2) {+}; \\
		\node[lmodule](decmlp) {MLP}; \\
		\coordinate (decres); \\
		\node[smodule](decnorm1) {Layer Norm}; \\
		\node[plus] (decplus1) {+}; \\
		\node[lmodule](decmha) {Multi-Head Attention}; \\
	};
	
	\begin{pgfonlayer}{background}
	\node[repeatbox,fit=(decnorm2)(decscene),label={left:2x}] (decblock) {};
	\end{pgfonlayer}
	
	\node[data, below=2.5cm of declin3,] (exprinpdec) {Expr $e_{D}$}; 
	\node[kpdata, left=1.4cm of exprinpdec] (kpinpdec) {KP $k_D$};
	\node[query,  minimum width=1.1cm] at ($(exprinpdec)!0.5!(kpinpdec)$) (querypix) {Query Pixel};
	\node[query, above=0.5cm of exprinpdec] (query) {Query Repr.};
	\node[query, draw=none, fill=none] (position) at ($(querypix)!0.5!(kpinpdec)$) {};
	\node[encmodule,  minimum width=1.1cm] at ($(position|-query)$) (posenc) {Posenc};	
	\node[mlp,minimum width=1.1cm, above=0.2cm of query] (initmlp) {   MLP   };

	\node[lmodule,align=center,anchor=south,minimum width=1cm] (decrender) at ($(decblock.north-|decnorm2)+(0,0.08)$) {Render MLP};
	
	\node[imgdata] at ($(xdrivingimage -| decrender)$) (decout) {Output Color};
	
	\node[text=darkgray] at ($(decout|-text)$) (textdec) {\sffamily\small Decoder};

	\coordinate (decr) at ($(decblock.east)+(-0.13,0)$);
	\draw[-] (kpinpdec.north)--($(kpinpdec.north)+(0,0.18)$) -|(posenc);
	\draw[-] (querypix.north)--($(querypix.north)+(0,0.18)$) -|(posenc);
	\draw[-latex] (exprinpdec.north)--(query);
	\draw[-latex] (posenc)--(query);
	\draw[-] (query.north)--(initmlp);
	\draw[-latex] ($(initmlp|-decscene)$)--($(initmlp|-decscene)+(0.55,0)$) |- (decplus1);
	\draw[-] (decmha) -- (decplus1);
	\draw[-] (decplus1) -- (decnorm1);
	\draw[-] (decnorm1) -- (decmlp);
	\draw[-] (decmlp) --(decplus2) ;
	\draw[-] (decplus2) -- (decnorm2);
	\draw[-] (decnorm2) -- (decrender);
	\draw[-latex] (decrender) -- (decout);
	\draw[-latex] (decres) -- (decres-|decr) |- (decplus2);

	\begin{pgfonlayer}{background}
	\draw[-latex] (declin1|-decscene.north) -- (declin1|-decmha.south);
	\draw[-latex] (declin2|-decscene.north) -- (declin2|-decmha.south);
	\draw[-latex] (declin3|-initmlp.north) -- (declin3|-decmha.south);
	\end{pgfonlayer}
	
	\end{scope}

	\begin{scope}
	\begin{pgfonlayer}{background}
	\draw[-latex,dashed, thick,gray] (latentsinput.west)--(cnninputexpr1);
	\draw[-,dashed, thick,gray] (latentsinput1)--(latentsinput);
	\draw[-latex,dashed, thick,gray] (kpssrc)--(cnninputkp1);
	\draw[-,dashed, thick,gray] (kpssrc1)--(kpssrc);

	\draw[-latex,dashed, thick,gray] (kpsdriv.south) -- ($(kpsdriv.south |- kpinpdec)+(0,-0.55)$) -| (kpinpdec.south);
	\draw[-latex,dashed, thick,gray] (latentsdiving.south) -- ($(latentsdiving.south|-exprinpdec)+(0,-0.55)$) -| (exprinpdec.south);
	\draw[-latex,dashed, thick, gray] (cnnpatchemb.east) -- ($(cnnpatchemb.east)+(0.85,0)$) |- (encpe2);
	\draw[-latex,dashed, thick, gray] (cnnpatchemb.east) -- ($(cnnpatchemb.east)+(0.85,0)$) |- (encpe);
	\draw[-latex,dashed, thick, gray] (encscene.east) -- ($(encscene.east)+(1.3,0)$) |- (decscene);
	\end{pgfonlayer}
	\end{scope}

	\end{tikzpicture}%
	\vspace{-1.5ex}
	\caption{Architecture details.  Given the driving frame and source images, we extract facial keypoints and latent expression vectors. Extracted source information are used to generate the input representation of the Patch CNN. The encoder infers the set-latent source face  representation from the patch embeddings as in SRT~\cite{srt}. The decoder is applied for each query pixel individually and is conditioned with the driving keypoints and the latent driving expression vector.  For further implementation details we refer to the Supplementary Material.}
	\label{fig:arch}
\end{figure*}

\section{Method}

The SRT architecture~\cite{srt} encodes one or more posed images to an internal representation and reconstructs views from arbitrary viewpoints.
We adapt the architecture and the input representation in such a way that we learn an internal representation from one or more unposed facial images. Reconstruction then allows free choice of head pose and facial expression.
Given a set-latent representation of an encoded face, head pose and facial expression can be controlled by ten keypoints and a latent expression vector. Abstractly, the internal representation can also be understood as an embedding that separates the appearance of a person from the head pose and expression.

Given an input representation $\{R_{S_i}\}$ (see \cref{sec:repr}), the transformer encoder $\mathcal{E}$ (\cref{fig:arch,sec:encoder}) produces a set-latent scene representation
\begin{equation}
\{z_z \in \mathbb{R}^d\} = \mathcal{E}(\text{CNN}(\{R_{S_i}\})),
\end{equation}
where CNN (\cref{sec:cnn}) is a convolutional feature extractor backbone (shared in case of multiple input images), as proposed by \cite{srt}.
Given this set-latent representation and the query $Q_{I_D}(q)$ (see \cref{sec:repr}), the transformer decoder $\mathcal{D}$ predicts the pixel color
\begin{equation}
C(q) = \mathcal{D}(Q_{I_D}(q) \; | \; \{z_z\}).
\end{equation}
Our full architecture is visualized in \cref{fig:arch}.

\subsection{Input and Query Representation}\label{sec:repr}
Given are one more facial source images $I_{S_i}$. We encode each image through ten keypoints $k_{S_i}$, computed by a keypoint detector, and a latent expression vector $e_{S_i}$ which we learn in self-supervised manner. The keypoints are normalized to $(-1,1)$ and positionally encoded~\cite{nerf}
\begin{alignat}{2}
f(p, s_O, O) &= \concat_{m=s_O}^{s_{O}+O-1} \sin(2^m\pi p) \oplus \cos(2^m\pi p)
\intertext{to obtain}
\gamma_\text{key}(k_{S_i}) &= \concat_{j=0}^{n_{\mathcal{K}}} f\left(k_{S_i}^{(j)}, s_{O_\text{key}}, O_\text{key}\right) \label{eq:positional}
\end{alignat}
where $n_{\mathcal{K}}$ is the number of keypoints, $O_{\text{key}}$ is the number of octaves per keypoint, $s_{O_\text{key}}$ is the keypoint start octave, and $\concat$ is the vector concatenation.

During face reenactment, keypoints might move out of the image boundaries $(-1,1)$.
Due to this, we set $s_{O_\text{key}}=-1$ to add an additional negative octave, which extends the interval of uniquely encodable values to $(-2,2)$. We generate training samples with keypoints outside the image by estimating them before cropping.

Similar to the keypoints, each image pixel with normalized coordinate $p=(x,y)$ receives a 2D positional encoding~\cite{nerf}
\begin{equation}
\gamma_\text{pix}(p,O_{\text{pix}}, s_{O_\text{pix}}) = f\left(p, s_{O_\text{pix}}, O_\text{pix} \right).
\end{equation}
The input representation of the source images $I_{S_i}$ at pixel $p=(x,y)$ is then given by
\begin{equation}
R_{S_i} (p) = [c_p,\gamma_\text{pix}(p),\gamma_\text{key}(k_{S_i}),e_{S_i}],
\end{equation}
where $c_p$ is the RGB color at pixel $p$ and $e_{S_i}$ is the latent expression vector
extracted from $I_{S_i}$ (see \cref{sec:expression}).
Note that each pixel is conditioned with the full keypoint encoding and the full latent expression vector, which quickly leads to a large input representation.

The decoder is queried for every output pixel $q=(x',y')$. Instead of the camera pose, as in \citep{srt}, each positionally-encoded query pixel is additionally conditioned with the desired target keypoints $k_D$ (i.e. the driving keypoints) and the latent expression vector $e_D$ of the desired expression (i.e. the latent expression vector of the driving frame).

Hence, the pixel-wise query representation is
\begin{equation}
Q_{I_D}(q) = [\gamma_\text{pix}(q), \gamma_\text{key}(k_{D}), e_{D}].
\end{equation}
Intuitively, the decoder attends to the most important features of the representation $\{z_z\}$ to render the source image appearance with the desired motion given by $k_D$ and $e_D$.

Note that our method does not require camera extrinsics or intrinsics,
since we operate directly in pixel space.

\paragraph{Keypoint Detector.}
The keypoint detector $\mathcal{K}$ is a fully convolutional hourglass network~\cite{hourglass}, as proposed by \citet{fom,monkey_net}.
For each input image $I$, it predicts heatmaps $H_I^{(i)} \in [0,1]^{H\times W}, \,\, i= 1,\dots,n_{\mathcal{K}}$, which define the pixelwise presence confidence of keypoint $k_i$. For all experiments, we fix the number of keypoints to $n_{\mathcal{K}}=10$.

\paragraph{Expression Network.}\label{sec:expression}
We assume that the last feature maps $F_{I,\mathcal{K}}$ predicted by the keypoint detector capture local image properties.
Given this assumption we build an expression network $\mathcal{X}$ that recycles $F_{I,\mathcal{K}}$ and the predicted keypoint heatmaps $H_I$ for an image $I$  (see \cref{fig:arch}).
We filter $F_{I,\mathcal{K}}$ by a learned $7\times7$ convolution $\mathcal{X}_C$, producing $F_{I,\mathcal{X}}$ with shape $[n_{f}, n_{\mathcal{K}}, h', w']$.
To focus on facial features, we utilize $H_I$ to aggregate the features spatially for each keypoint $k_i$:
\begin{equation}
{\vec{f}_{I,\mathcal{X}}}^{\,\,(i)} = \concat_{c=1}^{64} \left[\sum_{y=0}^{h'} \sum_{x=0}^{w'} H^{(i)}_I(x,y) F_{I,\mathcal{X}}^{(c)}(i,x,y)\right] \in \mathbb{R}^{n_{f}} \label{eq:8}
\end{equation}
to obtain
\begin{equation}
f_{I,\mathcal{X}} =  \concat_{i=1}^{n_{\mathcal{K}}} \left[ \vec{f}_{I,\mathcal{X}}^{\,\,(i)} \right]  \in \mathbb{R}^{n_{\mathcal{K}} \cdot n_{f}},
\end{equation}
where $c$ is the channel index, $i$ is the keypoint index, and $\concat$ is the concatenation operation.
Using a 4-layer MLP $\mathcal{X}_M$, we compute the latent expression vector
\begin{equation}
e_I = \mathcal{X}_M\left( \,f_{I,\mathcal{X}} \right).
\end{equation}
The expressional information of all important keypoints areas are thus spread throughout $e_I$.
Additionally, keypoint regions that do not contain important expression information can be filtered out by combining the local information.

\paragraph{Patch CNN.} \label{sec:cnn}
As in \citep{srt}, the shared CNN is designed to reduce the spatial dimension of the input data and fuse patch information. In each block, the height and width are reduced by factor two and the number of feature maps is doubled.
For all experiments, we use three CNN blocks followed by a final convolution with kernel size one, which generates the number of feature maps $n^{fm}_{\mathcal{E}}$ needed for the transformer encoder. The features with shape $[bs,n^{fm}_{\mathcal{E}},H/8,W/8]$ are then reshaped to $[bs,\frac{H+W}{8},n^{fm}_{\mathcal{E}}]$ which is the patch embedding input to the encoder.

\paragraph{Encoder.} \label{sec:encoder}
Following \cite{srt}, the standard transformer alternates global self-attention (between all tokens) and small MLP networks (see \cref{fig:arch}). Following \cite{osrt}, we drop source ID embeddings and reduce the number of attention blocks to five. Through self-attention across all patch embeddings, the encoder learns a set-latent scene representation $\{z_z\}$ in which each vector $z_z$ captures global scene and dynamics information.
Note that the cardinality of the set-latent representation scales linearly with the number of source images.

\paragraph{Decoder.}\label{sec:dec}
The transformer decoder does not use self-attention, but instead attends the set-latent scene representation with the query $Q_{I_D}$.
This is repeated for two times, followed by a render MLP that predicts the final output color at a certain pixel location.
For better performance, the query is first fed through a small 2-layer MLP that spreads the information in all dimensions, as proposed by \cite{osrt}.
Furthermore, we also use a final 5-layer render MLP that predicts the output color given the output of the attention module.
Intuitively, the decoder learns to attend to the most important features of $\{z_z\}$ to infer the pixel information of the encoded facial image with the requested head pose and facial expression.
Note that unlike SRT, we do not only request a novel view of a static scene but also certain dynamics within the scene, such as mouth movement.
It is therefore necessary for the encoder to output a highly flexible scene representation (see experiments with small decoder in \cref{exp:smallD}).

Due to the transformer design, the decoder can handle $\{z_z\}$ of any cardinality. Thus,
a trained encoder and decoder can operate on a flexible number of source images.

For a given source face, we only need to predict the set-latent scene representation once and each query pixel is estimated independently of the others. This is an advantage over CNN-based approaches~\cite{spline,dagan,fom,monkey_net,one-shot}, because it allows the model throughput to be scaled linearly with the number of available GPUs. Only one copy of the decoder needs to reside on each GPU.

\subsection{Augmentation and Regularization}\label{subsec:regularization}
Ideally, the network should learn to decouple appearance, pose, and expression information into set-latents $z_z$, keypoints $k_I$, and expression vector $e_I$, respectively.
This separation of concerns enables cross-reenactment.
In practice, the method is prone to overfitting, since we can only train in the self-reenactment regime, where ground truth is available.
This results in latents that jointly encode appearance, pose and expression,
which is visible when cross-reenacting to a different person.
Artifacts appear in the background area around the face and the model also deforms the source person to be closer to the face shape of the driving frame (see \cref{fig:reg}).
Hence, we do not reach the intended separation level.
To combat this, we implement several data augmentation and regularization measures.

\paragraph{Color Augmentation.}
To prevent colors leaking from the driving frame to the output image, we apply
color jitter augmentation on the source images.
Specifically, we create two color-jittered versions $I_S^{A1}, I_S^{A2}$ of the input image $I_S$.
The expression network $\mathcal{X}$ is run to extract expression vectors $e_{I_S^{A1}}, e_{I_S^{A2}}$.
An additional regularization term is added to enforce invariance to color jitter:
\begin{equation}
\mathcal{L}_{\text{aug}} = \frac{1}{|e|}\left\Vert e_{I_S^{A1}}-e_{I_S^{A2}}\right\Vert_2^2. \label{eq:Laug}
\end{equation}
While the encoder $\mathcal{E}$ is always trained with RGB colors from $I_S^{A1}$,
it receives the expression vector $e_{I_S^{A2}}$.
This further improves color invariance.

\paragraph{Cropping.} \label{par:crop}

To reduce background information in the output of $\mathcal{X}$, we further randomly crop
the driving frame (just for $\mathcal{X}$).
Here, we define $\Omega(\cdot)$, which selects a random crop with awareness of facial keypoints as proposed by~\cite{bulat}.
This crop is then scaled back to the original size, which can change the aspect ratio.
We add a loss term
\begin{equation}
\mathcal{L}_{\text{aug,D}} = \frac{1}{|e|}\left\Vert e_{I_D^{A}} - e_{\Omega \left(I_D^{A3} \right)} \right\Vert_2^2 \label{eq:LaugD}
\end{equation}
on the expression vectors of color-jittered versions $I_D^{A}, I_D^{A3}$, in which $A$ is either $A_1$ or $A_2$.
Adding $\Omega{(\cdot)}$ to the loss term encourages that $\mathcal{X}$ extracts scale-invariant expression information only from the face region.
Primarily, the expression vector of $\Omega \left(I_D^{A3} \right)$ is also utilized for decoding. 
However, in 25\% of cases, $e_{I_D^{A}}$ is selected, which employs the same color-jittering ($A_1$ or $A_2$) applied to the source images.

\paragraph{Statistical Regularization.}
Data augmentation alone is not enough to completely prevent head pose, expression, and appearance information from being jointly encoded (see \cref{fig:reg}).
We take inspiration from VICReg~\citep{vicreg}, a method for regularization of unsupervised feature learning based on invariance, variance, and covariance, but adapt it to encourage the focus on expression information.
Invariance against augmentations is already covered by \cref{eq:Laug,eq:LaugD}.

The covariance part aims to decorrelate along the feature dimension.
Intuitively, decorrelation encourages separation of expression from head pose, shape, and appearance and enables the network to drop non-expressional information (which is already encoded
in keypoints and scene representation $\{z_z\}$).
Given a batch of source and driving images concatenated in the batch dimension
\begin{align}
E=[&e_{S_1}^{(1)},\dots,e_{S_{(n_{src})}}^{(1)},e_D^{(1)},\nonumber\\
&\hspace{1.7cm}\vdots\\
&e_{S_1}^{(bs)},\dots,e_{S_{(n_{src})}}^{(bs)},e_D^{(bs)}] \nonumber
\end{align}
with shape $[(n_{src}+1) bs, c]$,
we estimate the covariance of the individual dimensions $\Cov(E)$.
The covariance loss is
\begin{equation}
\mathcal{L}^E_\text{Cov} = \frac{1}{c}(\underbrace{\sum_{i\neq j}[\Cov(E)]^2_{i,j}}_\text{off  diagonal}+\underbrace{\sum_{k}[\Cov(E)]^2_{k,k}}_\text{diagonal (variance)}). \label{eq:cov}
\end{equation}
In contrast to VICReg, we minimize the diagonal variance terms as well,
which represents an additional information bottleneck. In experiments, this regularization was helpful.

Since we train with supervision, there is no risk of ending up in a mode collapse,
so the batch-variance criterion of VICReg is not required.
Conversely, we found that encouraging variance \textit{along the feature dimension} of each vector with a hinge loss (penalizing vanishing features)
\begin{equation} %
\mathcal{L}^E_\text{Var} = \frac{1}{|E|}\sum_{e\in E}\max \left(0, \; 1-\sqrt{\Var(e)+\epsilon} \right), \label{eq:var}
\end{equation}
leads to better results and a more stable training. 
Finally, we define $\mathcal{L}_{\text{Cov}}$$=$$\mathcal{L}^{E_{1}}_{\text{Cov}}$$+$$\mathcal{L}^{E_{2}}_{\text{Cov}}$ and 
$\mathcal{L}_{\text{Var}}$$=$$\mathcal{L}^{E_{1}}_{\text{Var}}$$+$$\mathcal{L}^{E_{2}}_{\text{Var}}$, where $E_{1}$ and $E_{2}$ are the differently augmented variants of $E$.

\subsection{Training} \label{sec:training}

We use the VoxCeleb dataset~\citep{voxceleb} and prepare it as suggested by \citet{fom}.
It consists of $\sim$3000 videos from 419 different identities divided into a total of $\sim$17.000 utterances with a resolution of 256$\times$256.
During training, we sample $n_{src}$ source frames and one driving frame from the same video.
Keypoints are extracted using the detector network of \citep{fom}, which is not trained further.

We train the rest of our method in three distinct phases. In all phases, we apply the regularization
loss
\begin{equation}
\mathcal{L}_{\text{reg}} = \tfrac{\lambda_{\text{aug}}}{2}\left(\overline{\mathcal{L}_{\text{aug}}} + \overline{\mathcal{L}_{\text{aug,D}}} \right) + \lambda_{\text{Cov}}\mathcal{L}_{\text{Cov}}+ \lambda_{\text{Var}}\mathcal{L}_{\text{Var}},
\end{equation}
where $\overline{\mathcal{L}_{\text{aug}}}$ and $\overline{\mathcal{L}_{\text{aug,D}}}$ are the mean values of $\mathcal{L}_{\text{aug}}$ and $\mathcal{L}_{\text{aug,D}}$ calculated over the entire batch.
We start in Phase\,I with warm-up training, optimizing the MSE loss~\citep{srt}
\begin{equation}
\mathcal{L}_{\text{MSE}} = \mathbb{E}_{q \sim I_D} \Vert \mathcal{D}(q) - I_D(q) \Vert_2^2,
\end{equation}
where we approximate $\mathbb{E}_{q \sim I_D}$ with 4096 sampled pixels.

Using only pixel-wise losses leads to blurry images (see \cref{fig:reg}). To address this issue, we propose to add the perceptual loss $\mathcal{L}_{P}$ \citep{perceptual} in Phase\,II to generate more details.
During our experiments, we found that the batch size must be large enough to avoid local minima and poor performance. Since training on the full frames already exceeds 80GB
with a batch size of four, we compute gradients only sub-sampled to $128^2$ pixels
and compute the remaining pixels without gradient information. We apply a random pixel offset to ensure that all positions are covered during training.
This trick allows us to estimate image-based losses without requiring costly gradient estimation for the entire image.

Finally, in Phase\,III, we then add adversarial losses $\mathcal{L}_{A}$, which guide the model to predict realistic images. Following \citet{fom}, we utilize a CNN-based keypoint-aware discriminator $\mathcal{A}$ with 4 blocks
and also add a feature matching loss $\mathcal{L}_A^F$ between the discriminator maps predicted from the generated image and the ground truth image. %

The final loss in phase three is thus:
\begin{align}
\begin{split}
\mathcal{L} = \mathcal{L}_{\text{reg}} + \lambda_{\text{MSE}} \mathcal{L}_{\text{MSE}} + \lambda_{P} \mathcal{L}_{P}
&+ \lambda_A \mathcal{L}_{A} + \lambda_A^F\mathcal{L}_A^F.
\end{split}
\end{align}

In our experiments, we train with a batch size of 24 on three NVIDIA A100 GPUs (80GB), for 200k iterations in Phase\,I, 300k iterations in Phase\,II, and approximately 500k iterations in Phase\,III, depending on the validation performance (see Suppl. Material).
We set $\lambda_{\text{MSE}}=1$, $\lambda_P=0.01$, $\lambda_A=0.001$, $\lambda_A^F=0.01$, $\lambda_{\text{aug}}=\lambda_{\text{Cov}}=1$, and $\lambda_{\text{Var}}=0.2$. Especially Phase\,I is very important to avoid overfitting. When skipped, we experience extremely slow training progress and easily end up in a bad local minimum.

\subsection{Inference} \label{sec:infer}
For self-reenactment, the inference follows the training pipeline. In contrast,
for cross-reenactment, the driving frame comes from a different person.
This means that keypoints $k_D$ can be taken as-is (absolute motion transfer) or
adapted (relative motion transfer).
This adaption is calibrated from a selected driving frame that best matches the head pose and expression (measured through the normalized keypoints of \citet{bulat}).
Following \citet{fom}, the scale is estimated by comparing the volume of the convex hull
of head keypoints.
Driving keypoint movement is then scaled correctly and added to the keypoints of the source image.

In both cases, the facial expression vector does not depend on pose, shape, or appearance
and is applied as-is, which is a particular advantage of our method.

\section{Experiments} \label{sec:experiments}
In this section, we carry out various experiments on the official VoxCeleb test dataset~\cite{voxceleb} with image size 256$^2$. Additional results are reported in the Supplementary Material.

\subsection{Self-reenactment}
In self-reenactment, the source image is selected as the first frame in the driving video. In the case of two source images, we also select the last frame. We then reconstruct every tenth frame within the video, ensuring that each driving frame is at least ten frames apart from the closest source image.
We compare the animations to ground truth using the Peak Signal-to-Noise
Ratio (PSNR), Structural Similarity (SSIM)~\citep{ssim}, mean L1 error, and Average Keypoint Distance (AKD). To compute the AKD, we utilize external facial keypoints provided by \citet{bulat}.

In \cref{tab:selfre}, we compare with related methods. Our multi-source ablation outperforms related methods in terms of accuracy. For single source images, we achieve state-of-the-art performance. 
While TSMM~\cite{spline} slightly outperforms our method in SSIM, AKD, and PSNR, we note that they inpaint only disoccluded regions of the detected background. This produces nearly perfect reconstructions in static background areas that are also visible in the source image.
Furthermore, our method generalizes much better for cross-reenactment and produces more temporally consistent animations, as highlighted with our user study (see \cref{tab:study}).

While a low AKD and high SSIM value are good for self-reenactment, they often indicate that face shapes predicted by a model are highly dependent on the driving face structure. This, however, is detrimental for cross-reenactment, where the source appearance may be distorted by poorly matching driving keypoints (see shape deformations of related methods in \cref{fig:abs}). Also, for  relative motion, the alignment assumption (explained in \cref{sec:infer}) is often not perfectly satisfied, leading to poorly matching keypoints. Our method is more robust to this (see \cref{fig:rel}), because we do not use the structure of the driving frame to estimate the optical flow and we encode appearance information invariant to the driving keypoints in the set-latent scene representation.

\begin {table} \centering \footnotesize  %
\begin{ThreePartTable}
\begin{tabu} to \linewidth {lr@{\hspace{18pt}}rrrr}
\toprule

	Method                              & \#KP & SSIM$\uparrow$    & PSNR$\uparrow$     & L1$\downarrow$   & AKD$\downarrow$ \\

	\midrule
	DPE \cite{dpe}                      & 0 & .7180             & 22.94              & .0484            & 3.07  \\
	FOMM \cite{fom}                     & 10 & .7310             & 22.90              & .0470            & 2.26  \\
	DaGAN \cite{dagan}                  & 15 & .7563             & 23.51              & .0450            & 2.10  \\
	DaGAN$/$dv2 \cite{dagan}\tnote{1}   & 15 & .7346             & 22.81              & .0493            & 2.50  \\
	OSFS \cite{one-shot}\tnote{2}       & 15 & .7327             & 22.97              & .0471            & 2.33  \\
	TSMM \cite{spline}                  & 50 & \underline{.7660} & \underline{23.76}  & .0433            & \textbf{2.00}  \\
	\midrule
	Ours${/2\text{-Src}}$               & 10 & \textbf{.7891}    & \textbf{25.00}     & \textbf{.0360}   & \underline{2.04} \\
	Ours                                & 10 & .7576             & 23.67              & \underline{.0421}& 2.13  \\
	\hspace{0.5em} $|e|=128$                     & 10 & .7558             & 23.56              & .0428            & 2.16   \\
	\hspace{0.5em} $|e|=64$                      & 10 & .7535             & 23.61              & .0430            & 2.18   \\
	\hspace{0.5em} $\text{small}\,\mathcal{D}$  & 10 & .7548             & 23.60              & .0430            & 2.17   \\
	\hspace{0.5em} $n_{\mathcal{K}}=0$          & 0  & .7445             & 23.56              & .0436            & 2.64  \\
	\bottomrule
\end{tabu}
\vspace{-1.5ex}
\caption{Self-reenactment results (including ablations) on the official VoxCeleb test set~\cite{voxceleb}. Underlined values are the second best.}
\label{tab:selfre}
\begin{tablenotes}
  \item[1] Uses depth network trained on VoxCeleb2~\citep{chung18b} for inference
  \item[2] Third-party implementation \vspace{-0.65ex}
\end{tablenotes}
\end{ThreePartTable}
\end{table}

\subsection{Ablation Study}\label{sec:ablation_study}
We run an ablation study to compare quantitative results (see \cref{tab:selfre}). A qualitative comparison and implementation details are reported in the Supplementary Material.

\paragraph{Do we need keypoints?}
We report an ablation without keypoint encoding~(Ours$/$$n_{\mathcal{K}}=0$), i.e. all pose
information is carried implicitly in the latent vector $e$, removing factorization of pose and expression,
which makes relative motion transfer impossible.
As can be seen in \cref{tab:selfre}, this change results in worse self-reenactment performance.
See Suppl. Material for qualitative cross-reenactment comparisons.

\begin{figure}\sf \tiny \centering
	\vspace{-0.5ex}
	\begin{tikzpicture}[
	spy using outlines={rectangle, magnification=2.9, size=1cm}
	]
	
	\node[inner sep=0] (img1)
	{\includegraphics{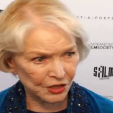}};
	\spy [red] on (rel cs:x=70,y=32,name=img1) in node [above] at ($(img1.north)+(0.45,-0.6)$);
	
	\node[below] at (img1.south) {Source};

	\node[inner sep=0,right=1pt of img1] (img1-2)
	{\includegraphics{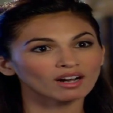}};

	\node[below] at (img1-2.south) {Driving};

	\node[inner sep=0,right=4pt of img1-2] (img2)
	{\includegraphics{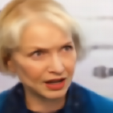}};
	\spy [red] on (rel cs:x=70,y=32,name=img2) in node [above] at ($(img2.north)+(0.45,-0.6)$);
	
	\node[below] at (img2.south) {Ours};
	
	\node[inner sep=0,right=1pt of img2] (img3)
	{\includegraphics{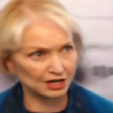}};
	\spy [red] on (rel cs:x=70,y=32,name=img3) in node [above] at ($(img3.north)+(0.45,-0.6)$);
	
	\node[below] at (img3.south) {w$/$o Stat. Reg.};

	\end{tikzpicture}
	\begin{tikzpicture}[
	spy using outlines={rectangle, magnification=1.1, size=1.3cm}
	]
	
	\node[inner sep=0] (img1)
	{\includegraphics{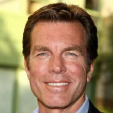}};
	\node[below] at (img1.south) {Source};
	
	\node[inner sep=0,right=1pt of img1] (img1-2)
	{\includegraphics{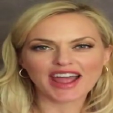}};
	\node[below] at (img1-2.south) {Driving};

	\node[inner sep=0,right=4pt of img1-2] (img2)
	{\includegraphics{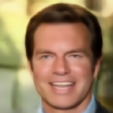}};
	\node[below] at (img2.south) {Ours};

	\node[inner sep=0,right=1pt of img2] (img3)
	{\includegraphics{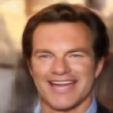}};

	\node[below] at (img3.south) {w$/$o Reg.};
	
	\end{tikzpicture}
	\vspace{-3ex}
	\caption{%
	Regularization benefit in Phase I training (relative motion transfer).
	If trained without statistical regularization (w$/$o Stat. Reg.), artifacts originating from the driving frame
	are visible in the background around the face boundary.
	When dropping regularization entirely (w$/$o Reg.), color distortions, background artifacts, and shape
	deformations are clearly visible.
	The lower sequence uses a source image from the CelebA-HQ dataset~\citep{celebahq}.
	}
	\label{fig:reg} \vspace{-2ex}
\end{figure}
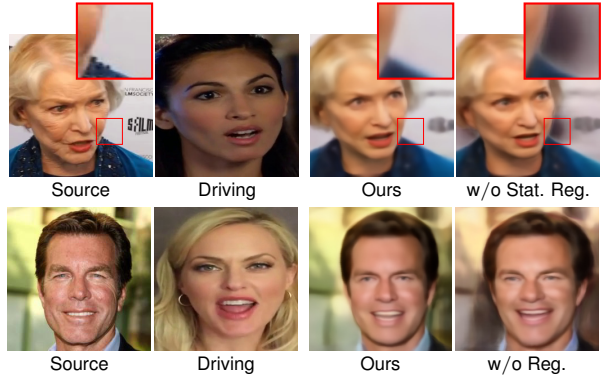

\paragraph{What size should latent vectors have?}
Our reference model is trained with one source image and a latent expression vector of size $|e|$=256.
As mentioned in \cref{subsec:regularization} and visualized in \cref{fig:reg}, training without our proposed regularization leads to overfitting, resulting in shape deformation, color distortion, and background artifacts.
With $|e|$=128 and $|e|$=64 latent expression dimensions, the self-reenactment performance decreases only slightly, showing that we can significantly reduce the amount of driving information being transmitted (e.g. for low-bandwidth videoconferencing) without losing much accuracy. In cross-reenactment, we also noticed a slight degradation in the transmission of facial expressions. %
However, the results are still good.

\paragraph{How efficient is the set-latent representation for decoding?} \label{exp:smallD}
We train a model (Ours$/\text{small}\,\mathcal{D}$) with a significantly smaller decoder (see Supplementary Material).
Interestingly, we achieve a self-reenactment performance close to the reference model. This indicates that the set-latent representation is very efficient to decode and already models facial dynamics. 
However, the sharpness of fine details, such as hair, is slightly degraded.
Reducing the decoder capacity increases throughput from 11\,fps to 23\,fps, enabling real-time application on a single NVIDIA RTX 4090 GPU. 
As mentioned in \cref{sec:dec}, we can scale throughput linearly with the number of GPUs without introducing additional latency.

\paragraph{Can we improve results with multiple source images?}
Unlike state-of-the-art methods~\cite{dpe,fom,dagan,spline,one-shot}, our architecture allows the use of an arbitrary and flexible number of source images when available (e.g., when extracted from a video).
Multiple source images can help the model understand person-specific face dynamics.
In this experiment, we train a model ablation (Ours$/$2-Src) with two source images.
As \cref{tab:selfre} shows, the results are significantly improved. Interestingly, the model generalizes to more than two source images even without explicit training.

\subsection{Cross-reenactment}
Our main motivation is to perform cross-reenactment. 
We sample 20 source images and driving videos from the official VoxCeleb~\cite{voxceleb} test set and compare our videos to state-of-the-art animations in a pairwise user study in \cref{tab:study}.
To be fair, we only use a single source image.
We also present qualitative results in \cref{fig:rel,fig:abs} and report additional user study information in the Supplementary Material.
In general, our method is better at cross-ID motion transfer, while producing more consistent and natural results.
For additional qualitative results on CelebV~\citep{wu2018reenactgan}, CelebA-HQ~\citep{celebahq}, and
VoxCeleb2~\citep{chung18b} see our Supplementary Material.
\newlength{\qqheight}\setlength{\qqheight}{1.35cm}
\newcommand{\imgqq}[1]{\includegraphics[height=\qqheight,clip,trim=0 0 0 0]{#1}}
\begin{figure}
\centering
\begin{tikzpicture}[
font=\sffamily\scriptsize,
a/.style={inner sep=0pt},
lu/.style={anchor=south,text=black,inner sep=0pt},
lu2/.style={anchor=south,text=black,inner sep=0.7pt},
ld/.style={anchor=north,text=black,inner sep=2pt}
]
\matrix (masks) [inner sep=0pt, matrix of nodes, every node/.style={a, inner sep=0pt}, column sep=0pt, row sep=0pt]
{
	|(m0)| \includegraphics{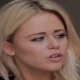}&
	|(m1)| \includegraphics{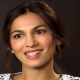}\hspace{0pt}&[1pt]
	|(m2)| \includegraphics{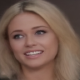}&
	|(m3)| \includegraphics{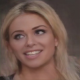}&
	|(m4)| \includegraphics{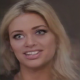}&
	|(m5)| \includegraphics{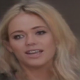}\\
	|(m7)| \includegraphics{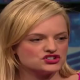} &
	|(m8)| \includegraphics{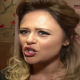}\hspace{0pt} &
	|(m9)| \includegraphics{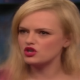} &
	|(m10)| \includegraphics{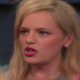} &
	|(m11)| \includegraphics{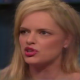} &
	|(m12)| \includegraphics{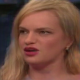} \\
	\includegraphics{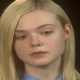} &
    \includegraphics{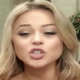}\hspace{0pt} &
    \includegraphics{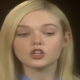} &
    \includegraphics{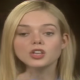} &
    \includegraphics{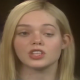} &
    \includegraphics{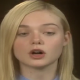} \\
    \includegraphics{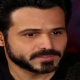} &
    \includegraphics{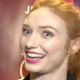}\hspace{0pt} &
    \includegraphics{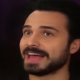} &
    \includegraphics{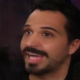} &
    \includegraphics{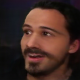} &
    \includegraphics{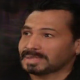} \\
    \includegraphics{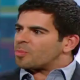} &
	\includegraphics{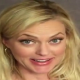}\hspace{0pt} &
	\includegraphics{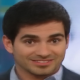} &
	\includegraphics{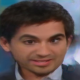} &
	\includegraphics{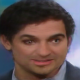} &
	\includegraphics{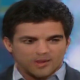} \\
};

\node[lu2] at (m0.north) {\raisebox{0mm}[0mm][0.6mm]{Source}};
\node[lu] at (m1.north) {\raisebox{0mm}[0mm][0.85mm]{Driving}};
\node[lu2] at (m2.north) {\raisebox{0mm}[0mm][0.6mm]{Ours}};
\node[lu] at (m3.north) {\raisebox{0mm}[0mm][0.75mm]{TSMM~\cite{spline}}};
\node[lu] at (m4.north) {\raisebox{0mm}[0mm][0.75mm]{DaGAN~\cite{dagan}}};
\node[lu] at (m5.north) {\raisebox{0mm}[0mm][0.75mm]{OSFS~\cite{one-shot}}};

\end{tikzpicture}%
\vspace{-1.5ex}
\caption{Cross-reenactment comparison with absolute motion transfer on the VoxCeleb test set~\cite{voxceleb}. We generate more accurate expressions with less shape deformations (higher ID preservation).}
\label{fig:abs} \vspace{-1ex}
\end{figure}

\begin {table} \centering \footnotesize  \setlength{\tabcolsep}{3pt}

\scriptsize
\begin{tabular}{lcccccc} \toprule

     & FOMM\cite{fom}     &   DaGAN\cite{dagan}  & \cite{dagan}$/$dv2  & TSMM\cite{spline} & OSFS\cite{one-shot} & DPE\cite{dpe}\\ \midrule
Rel.          & 97\% (20)             & 98\% (20)    & 95\% (19)         & 97\% (20)   & 87\% (19)  &  \\
Abs.           & 94\% (20)               & 99\% (20) & 96\% (20)         & 92\% (19)   & 94\% (19) & 95\% (20) \\
\bottomrule
\end{tabular}
\vspace{-2ex}
\caption{Cross-reenactment user study. Pairwise preferences between SOTA and our method. Higher values show higher preference for our videos. DPE~\cite{dpe} has no relative mode. ($\cdot$) shows the amount out of 20 scenes for which we got the majority of votes.}
\vspace{-1.7ex}
\label{tab:study}
\end{table}

\begin{figure}
	\centering
	\begin{tikzpicture}[
	font=\sffamily\scriptsize,
	a/.style={inner sep=0pt},
	lu/.style={anchor=south,text=black,inner sep=0pt},
	lu2/.style={anchor=south,text=black,inner sep=0.7pt},
	ld/.style={anchor=north,text=black,inner sep=2pt}
	]
	\matrix (masks) [inner sep=0pt, matrix of nodes, every node/.style={a, inner sep=0pt}, column sep=0pt, row sep=0pt]
	{
		|(m0)| \includegraphics{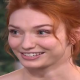}  &
		|(m1)| \includegraphics{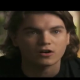}\hspace{0pt} &[1pt]
		|(m2)| \includegraphics{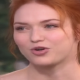}  &
		|(m3)| \includegraphics{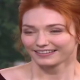} &
		|(m4)| \includegraphics{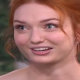} &
		|(m5)| \includegraphics{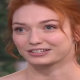} \\
		|(m7)| \includegraphics{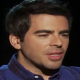} &
		|(m8)| \includegraphics{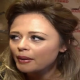}\hspace{0pt} &
		|(m9)| \includegraphics{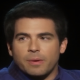} &
		|(m10)| \includegraphics{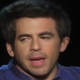} &
		|(m11)| \includegraphics{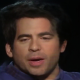} &
		|(m12)| \includegraphics{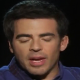} \\[0.15cm]
		\includegraphics{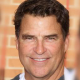} &
		\includegraphics{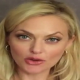}\hspace{0pt} &
		\includegraphics{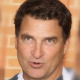} &
		\includegraphics{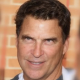} &
		\includegraphics{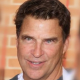} 
		&
		\includegraphics{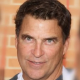} 
		\\
		\includegraphics{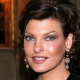} &
		\includegraphics{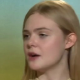}\hspace{0pt} &
		\includegraphics{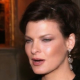} &
		\includegraphics{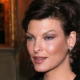} &
		\includegraphics{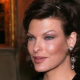} 
		&
		\includegraphics{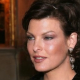} 
		\\
	};
	
	\node[lu2] at (m0.north) {\raisebox{0mm}[0mm][0.6mm]{Source}};
\node[lu] at (m1.north) {\raisebox{0mm}[0mm][0.85mm]{Driving}};
\node[lu2] at (m2.north) {\raisebox{0mm}[0mm][0.6mm]{Ours}};
\node[lu] at (m3.north) {\raisebox{0mm}[0mm][0.75mm]{TSMM~\cite{spline}}};
\node[lu] at (m4.north) {\raisebox{0mm}[0mm][0.75mm]{DaGAN~\cite{dagan}}};
\node[lu] at (m5.north) {\raisebox{0mm}[0mm][0.75mm]{OSFS~\cite{one-shot}}};
	
	\end{tikzpicture}%
	\vspace{-1.5ex}
	\caption{Comparison with SOTA in cross-reenactment with relative motion transfer.
	Our method is more robust to the alignment assumption for relative motion transfer, generates more accurate expressions, and handles larger pose offsets.
	All images are from the VoxCeleb test set~\citep{voxceleb}, except the lower block, which shows generalization to source images from the CelebA-HQ dataset~\citep{celebahq}.
	}
	\label{fig:rel}%
	\vspace{-1ex}
\end{figure}

\paragraph{Absolute Motion.}
When the driving keypoints are simply copied, users mainly prefer the animations generated by our method (see \cref{tab:study}). 
Since our method is more robust to poorly matching keypoints, we produce fewer shape deformations than other keypoint-based methods (see \cref{fig:abs}). Furthermore, we consistently animate larger pose offsets.

\paragraph{Relative Motion.}
More interesting and challenging is animating with relative motion. Here, best performance can be achieved when the facial expression representation is decoupled from head pose and shape.
As \cref{tab:study} illustrates, we significantly outperform state-of-the-art methods. %
When analyzing the results, we noticed that related methods show poor performance
when there is no good match for the source expression and head pose in the driving video.

\section{Limitations}
Our method struggles to generate out-of-distribution expressions such as sticking out the tongue or looking back.
While we produce sharper mouth and eye regions, details in the background and hair are sometimes slightly reduced, compared to CNN-based methods.
We believe that the model allocates most of its capacity to the face.
Compared to CNN approaches that simply learn to forward background pixels from the input,
our model must encode the background in the set-latents and reconstruct it by attending the correct features.
Increasing model capacity or optimizing the query representation might lead to improvements.%

\section{Conclusion}
We have proposed a state-of-the-art method for face reenactment.
To our knowledge, this is the first transformer-based architecture for this purpose.
We learn latent expression features that are free of appearance, shape or pose information, making them perfectly suited for cross-reenactment.
Our method achieves fast inference speed, which allows real-time application.
We proposed a regularization and training scheme which are necessary to guide the network to represent the scene as desired.
Future work could investigate further improving the animation quality of fine details (e.g. in the hair) and utilizing volume rendering techniques to reconstruct geometry.

{
    \small
    \bibliographystyle{ieeenat_fullname}
    \bibliography{main}
}

\clearpage
\setcounter{page}{1}
\maketitlesupplementary

\section{Implementation Details}
We present important training and architecture details, including the parameter values that were used.
\subsection{Architecture Details}
\paragraph{Keypoint Detector.}
The keypoint detector is used as-is from \citep{fom}
and not trained further. It consists of a 5-block Hourglass network~\cite{hourglass} with a block expansion of 32 and a maximum feature map size of 1024. For keypoint extraction, the images are resized to 64$\times$64. After decoding, the heatmaps are predicted by a final 7$\times$7 convolution. Keypoint locations are given by the centroids of the corresponding heatmap.

\paragraph{Latent Expression Extractor.}
The latent expression extractor $\mathcal{X}$ has a single 7$\times$7 convolutional layer that predicts $n_f=32$ individual feature maps for each keypoint. For each keypoint, the individual feature maps computed by the keypoint detector are aggregated in x and y direction with the weights of the corresponding heatmap. After aggregating the features of each keypoint individually, the information is concatenated and fused to predict a global expression vector. The fusion is performed by a 4-layer MLP with $(640-1280-640)$ hidden units and $|e|$ output neurons.

\paragraph{Input and Query Representation.}
For both, the positional encoding in the input and query representation, we set the number of octaves to $\mathcal{O}_{pix} = 16$ and $\mathcal{O}_{key} = 4$ with start octaves $s_{\mathcal{O}_{pix}} = -1$ and $s_{\mathcal{O}_{key}} = -1$. Together with a latent expression dimension of  $|e|=256$, this results in a query representation of size $|Q_{I_D}| = 416$ and an input representation $R_{S_i}$ with 419 input channels, since we also encode the RGB pixel color of the source image.

\paragraph{Patch CNN.}
In all experiments, we set the output feature dimension of the Patch CNN to $n^{fm}_{\mathcal{E}} = 768$. Since we are processing a very large number of input channels (419 when $|e|=256$), we use a bottleneck of 96 feature maps in the first convolutional layer.

\paragraph{Encoder.}

The transformer encoder also has a feature dimension of 768. Each multi-head attention layer uses 12 heads with an attention dimension of 64. The encoder processes the patch embedding of each source image individually, so that the cardinality of the set-latent scene representation scales linearly with the number of source images. This allows a flexible number of source images to be used. 
In total, the encoder and Patch CNN (with $|e|=256$) have $29$,$774$,$112$ parameters.

\paragraph{Decoder.}
The decoder has a feature dimension equal to the size of the query representation $|Q_{I_D}|$. 
The input MLP (see decoder in \cref{fig:arch})  has two layers with 720 hidden units and $|Q_{I_D}|$ output neurons. In the attention blocks, we also use 12 heads with an attention dimension of 64. The MLP inside the attention block, which fuses the information from the individual heads, has two layers and $2|Q_{I_D}|$ hidden units. The final 5-layer render MLP has $(1536-1536-1536-768)$ hidden units and three output neurons for the RGB color.

For our small decoder ablation Ours$/$small$\mathcal{D}$, we reduce the number of heads from 12 to 6 and also halve the number of hidden units of the MLP inside the attention block. Finally, we replace the render MLP with a smaller 3-layer version with $(1536-768)$ hidden units.
Compared to our standard decoder, the number of parameters is reduced from $15$,$310$,$131$ to $6$,$012$,$723$.

\paragraph{Discriminator.}
For the keypoint-aware discriminator $\mathcal{A}$, we use the implementation of \citet{fom} which is based on~\cite{pix2pix}.
The input is an RGB image concatenated with ten heatmaps representing the driving keypoints. In total, we use four blocks, resulting in 512 output features with a downsampling factor of 16. For further implementation and loss details, we refer to \citet{fom}.

\subsection{Training Details}
We train on three NVIDIA A100 (80GB) GPUs for about 23 days.
We found that warming up (i.e. Phase\,I training, explained in \cref{sec:training}) is essential to avoid ending up in local minima. Also, the batch size should be large enough. In our experiments we found out that 24 is sufficient. With a batch size of eight, training progressed slowly and appeared to be very unstable. Furthermore, we ended up in a local minimum with poor inference performance. When adding adversarial losses in training Phase\,III, we allow the discriminator to warm up for 500 iterations without computing gradients for the model. This is essential since otherwise the untrained discriminator will influence the current training progress with gradients of large magnitude.

 \paragraph{Stopping Criterion.}
We extract a validation dataset, which we use to validate the self- and cross-reenactment performance.
The self-reenactment performance is measured as in \cref{tab:selfre}. For cross-reenactment, we randomly sample source images and driving videos. Model performance is judged visually by us.
We found that it is not necessary to choose between good self- and cross-reenactment performance, as both are typically correlated.
We thus use self-reenactment scores as a way to find promising models and then verify cross-reenactment performance.

\paragraph{Visualizing Out-of-frame Motion.}
As explained in \cref{sec:repr}, we use a negative octave in the positional encoding of pixels and keypoints to uniquely encode values in $(-2,2)$. However, the VoxCeleb dataset~\cite{voxceleb} (prepared as suggested by \citet{fom}) itself has no out-of-frame motion. Instead, we create out-of-frame motion by cropping the image with respect to the source image keypoints. We use external pre-estimated face keypoints~\cite{bulat} and select a random crop of all selected images (source and driving) such that all source keypoints are inside. Finally, the images are resized back to 256$\times$256, which may change the aspect ratio and induces additional regularization. In some cases, the driving face will now be partially outside the image---generating corresponding training samples. 

Since cropping will reduce the image resolution to less than 256$^2$, we download the dataset at the highest resolution possible so that the crop (before resizing) is ideally larger than 256$^2$ and no image detail is lost. 

The keypoint detector can only predict keypoints within the image. Therefore, we detect keypoints of the uncropped images and use the cropping information to transform them into the cropped images. 

\begin{figure}\sf \tiny \centering
	\vspace{-0.5ex}
	\begin{tikzpicture}[
	spy using outlines={rectangle, magnification=3.5, size=1.3cm}
	]
	
	\node[inner sep=0] (img1)
	{\includegraphics{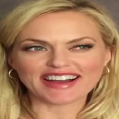}};

	\node[above] at (img1.north) {\raisebox{0mm}[0mm][0.6mm]Source};
	
	\node[inner sep=0,right=0pt of img1] (img1-2)
	{\includegraphics{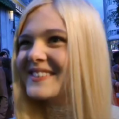}};

	\node[above] at (img1-2.north) {Driving};
	
	\node[inner sep=0,right=0pt of img1-2] (img2)
	{\includegraphics{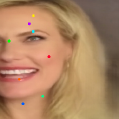}};

	\node[above] at (img2.north) {\raisebox{0mm}[0mm][0.6mm]Ours};
	
	\node[inner sep=0,right=0pt of img2] (img3)
	{\includegraphics{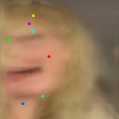}};

	\node[above] at (img3.north) {w$/$o Neg. Octave};

	\end{tikzpicture}

	\begin{tikzpicture}[
	spy using outlines={rectangle, magnification=4, size=1.3cm}
	]
	
	\node[inner sep=0] (img1)
	{\includegraphics{graphicscache/51818BD822F8EC13BF4BD247F5F45FFF.pdf}};

	\node[inner sep=0,right=0pt of img1] (img1-2)
	{\includegraphics{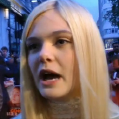}};

	\node[inner sep=0,right=0pt of img1-2] (img2)
	{\includegraphics{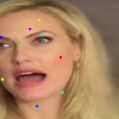}};

	\node[inner sep=0,right=0pt of img2] (img3)
	{\includegraphics{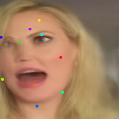}};

	\end{tikzpicture}

	\begin{tikzpicture}[
	spy using outlines={rectangle, magnification=4, size=1.3cm}
	]
	
	\node[inner sep=0] (img1)
	{\includegraphics{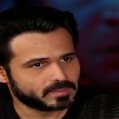}};

	\node[inner sep=0,right=0pt of img1] (img1-2)
	{\includegraphics{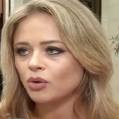}};

	\node[inner sep=0,right=0pt of img1-2] (img2)
	{\includegraphics{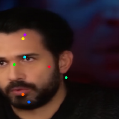}};

	\node[inner sep=0,right=0pt of img2] (img3)
	{\includegraphics{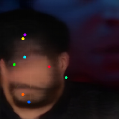}};

	\end{tikzpicture}

	\begin{tikzpicture}[
	spy using outlines={rectangle, magnification=4, size=1.3cm}
	]
	
	\node[inner sep=0] (img1)
	{\includegraphics{graphicscache/53415F1AC3EBB893D870652B50F3350A.pdf}};

	\node[inner sep=0,right=0pt of img1] (img1-2)
	{\includegraphics{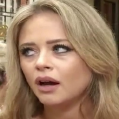}};

	\node[inner sep=0,right=0pt of img1-2] (img2)
	{\includegraphics{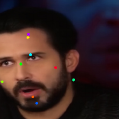}};

	\node[inner sep=0,right=0pt of img2] (img3)
	{\includegraphics{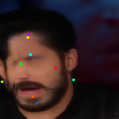}};

	\end{tikzpicture}

	\caption{Out-of-frame motion with (Ours) and without explicit addressing keypoints outside the image (w$/$o Neg. Octave). Out-of-frame motion only occurs when relative motion transfer is used (see \cref{sec:infer}). The predicted images are visualized with the driving keypoints that were used in the decoder. Images from the VoxCeleb test set~\citep{voxceleb}.}
	\label{fig:oof} 
\end{figure}

Unlike the source keypoints, the latent expression vectors are extracted directly from the cropped source images. When extracting expression vectors from the driving frame, the differently augmented driving frame version (as explained in \cref{par:crop}), ensures that the driving face is inside the image. 
In \cref{fig:oof}, we show that not addressing out-of-frame motion leads to poor results when keypoints are outside the image or close to the image boundaries.

\subsection{User Study Details}
We selected 30 different people to participate in the user study (see \cref{tab:study}). Since we compared the methods in pairs, each participant was only allowed to judge one related method. Furthermore, each participant judged both relative motion transfer and absolute motion transfer.
The face reenactment task was initially explained, and participants were instructed to base their decision on the following two criteria:
\begin{enumerate}
\item Does the motion transfer work well (including ID preservation)?
\item Does the animation look like a natural and consistent video?
\end{enumerate}
Each participant was simultaneously shown the source image, the driving video, our result and the animation of the comparison method. In each of the 20 sequences, we randomized whether our method was shown on the left or on the right. Participants could only decide once the video had run through. However, the video automatically restarted, so that there was no overall time limit. A decision was made by clicking on the preferred video.

\begin {table} \centering \small  \setlength{\tabcolsep}{8pt}
\begin{tabular}{@{}l@{\hspace{15pt}}ccccrr@{\hspace{0pt}}r@{}} \toprule

	Method                             & SSIM$\uparrow$    & PSNR$\uparrow$     & L1$\downarrow$   & AKD$\downarrow$ \\

	\midrule
	Ours                                & .7576             & 23.67              & .0421            & 2.13   \\
	Ours${/\,1\rightarrow2\text{-Src}}$ & .7181             & 23.06              & .0453            & 2.42  \\ \midrule
	Ours${/\,2\text{-Src}}$             & .7891             & 25.00              & .0360            & 2.04  \\
	Ours${/\,2\rightarrow3\text{-Src}}$ & .8092             & 25.80              & .0325            & 2.00 \\
	Ours${/\,2\rightarrow1\text{-Src}}$ & .7610             & 23.85              & .0418            & 2.13 \\

	\bottomrule
	\end{tabular}%
	
\vspace*{1mm}\raggedright 
\footnotesize 
~~~\mbox{Ours$/t\rightarrow i$-Src} means that the model trained with $t$ source images is\\
~~~evaluated with $i$ source images during inference.

\caption{Self-reenactment results on the official VoxCeleb test set~\cite{voxceleb} when generalizing to a different number of source images without explicit training. Training with two source images increases self-reenactment performance, even when only one source image is used for inference.}
\label{tab:addrelselfre}
\end{table}

\section{Additional Experiments \& Results}

We report auxiliary experiments and more qualitative results here.

\subsection{Flexibility in the Number of Source Images}

We investigate the generalization behavior with respect to changing the number of source images during inference. Here, our reference model was trained with a single source image and with two source images. 
As reported in \cref{tab:addrelselfre}, the model trained with two source images generalizes in both directions, with fewer and with more source images used for inference. Interestingly, when reducing the number of source images to one (line \mbox{Ours$/2\rightarrow 1$-Src} in \cref{tab:addrelselfre}) it even produces slightly better self-reenactment results than our model explicitly trained with only one source image (line \mbox{Ours} in \cref{tab:addrelselfre}). With three source images available for inference (line \mbox{Ours$/2\rightarrow 3$-Src} in \cref{tab:addrelselfre}), the performance increases further, indicating that additional source images can be added at inference as available.

The model trained with only one source image shows a significant drop in performance when the number of source images is increased during evaluation (line \mbox{Ours$/1\rightarrow 2$-Src} in \cref{tab:addrelselfre}). 
Therefore, if a flexible number of source images is desired, we recommend training with at least two source images. Alternatively, the number of source images can be chosen flexibly during training. To ensure that the data can still be batched, we recommend always selecting the maximum number of source images, but masking the set-latents of unnecessary source images in the attention module of the decoder.

\subsection{Ablation Study}
\begin{figure*}
	\centering
	\begin{tikzpicture}[
	font=\sffamily\small,
	a/.style={inner sep=0.1pt},
	lu/.style={anchor=south,text=black,inner sep=2pt},
	ld/.style={anchor=north,text=black,inner sep=2pt},
	img/.style={inner sep=0pt},
	]
	\matrix (masks) [inner sep=0pt, matrix of nodes, every node/.style={a, inner sep=0pt}, column sep=0pt, row sep=0pt]
	{
		|(m0)| \includegraphics{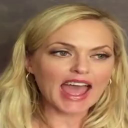}  &
		|(m1)| \includegraphics{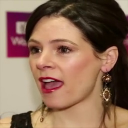}\hspace{1pt} &
		|(m2)| \includegraphics{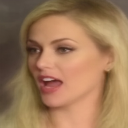}  &
		|(m3)| \includegraphics{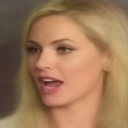} &
		|(m64)| \includegraphics{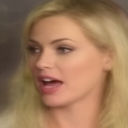} &
		|(m4)| \includegraphics{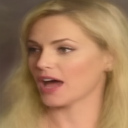} &
		|(m5)| \includegraphics{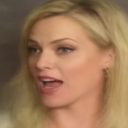}\hspace{1pt} & 
		|(m6)| \includegraphics{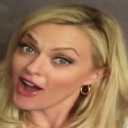} \\
		|(m7)| \includegraphics{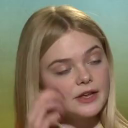} &
		|(m8)| \includegraphics{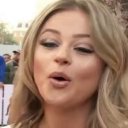}\hspace{1pt} &
		|(m9)| \includegraphics{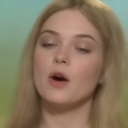} &
		|(m10)| \includegraphics{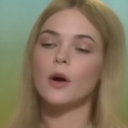} &
		|(m64_2)| \includegraphics{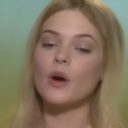} &
		|(m11)| \includegraphics{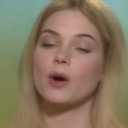} &
		|(m11)| \includegraphics{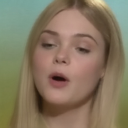}\hspace{1pt} &
		|(m12)| \includegraphics{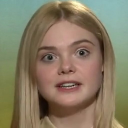} \\
		\includegraphics{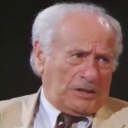} &
		\includegraphics{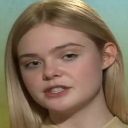}\hspace{1pt} &
		\includegraphics{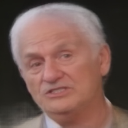} &
		\includegraphics{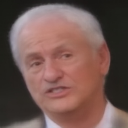} &
		\includegraphics{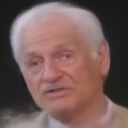} &
		\includegraphics{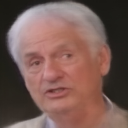} &
		\includegraphics{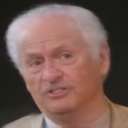}\hspace{1pt} &
		\includegraphics{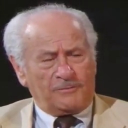} \\
		\includegraphics{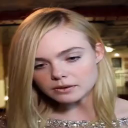} &
		\includegraphics{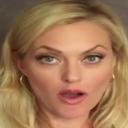}\hspace{1pt} &
		\includegraphics{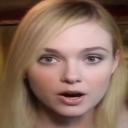} &
		\includegraphics{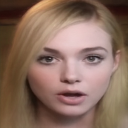} &
		\includegraphics{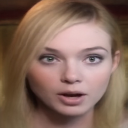} &
		\includegraphics{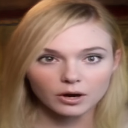} &
		\includegraphics{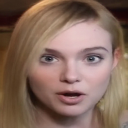}\hspace{1pt} &
		\includegraphics{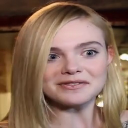} &	
		& \node[minimum height = 8.5pt] {};
		& 
		& 
		& 
		& 
		&  \\

		\includegraphics{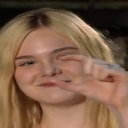} &
		\includegraphics{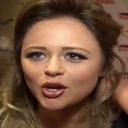}\hspace{1pt} &
		\includegraphics{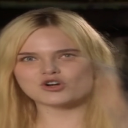} &
		\includegraphics{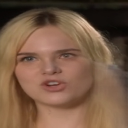} &
		\includegraphics{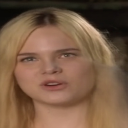} &
		\includegraphics{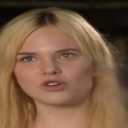} &
		\includegraphics{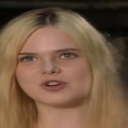}\hspace{1pt} &
		\includegraphics{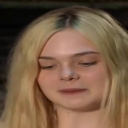} \\
		\includegraphics{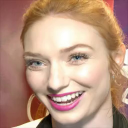} &
		\includegraphics{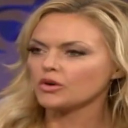}\hspace{1pt} &
		\includegraphics{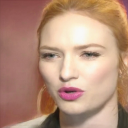} &
		\includegraphics{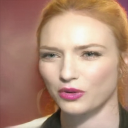} &
		\includegraphics{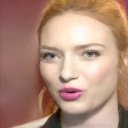} &
		\includegraphics{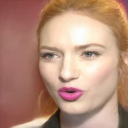} &
		\includegraphics{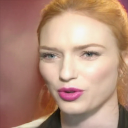}\hspace{1pt} &
		\includegraphics{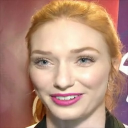} \\
		\includegraphics{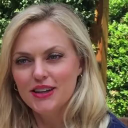} &
		\includegraphics{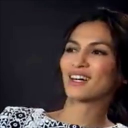}\hspace{1pt} &
		\includegraphics{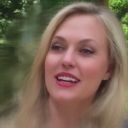} &
		\includegraphics{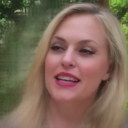} &
		\includegraphics{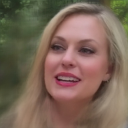} &
		\includegraphics{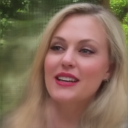} &
		\includegraphics{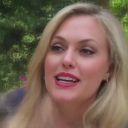}\hspace{1pt} &
		\includegraphics{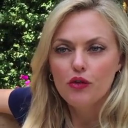} \\
		\includegraphics{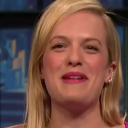} &
		\includegraphics{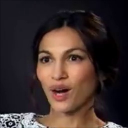}\hspace{1pt} &
		\includegraphics{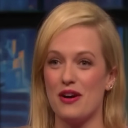} &
		\includegraphics{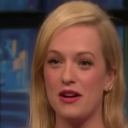} &
		\includegraphics{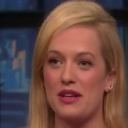} &
		\includegraphics{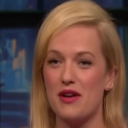} &
		\includegraphics{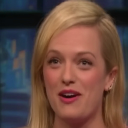}\hspace{1pt} &
		\includegraphics{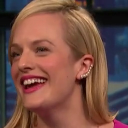} \\
		\includegraphics{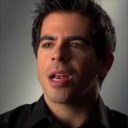} &
		\includegraphics{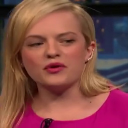}\hspace{1pt} &
		\includegraphics{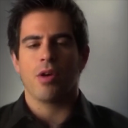} &
		\includegraphics{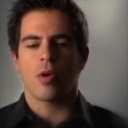} &
		\includegraphics{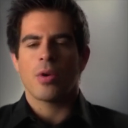} &
		\includegraphics{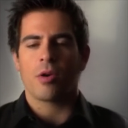} &
		\includegraphics{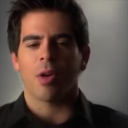}\hspace{1pt} &
		\includegraphics{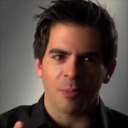} \\
	};
	
	\node[lu] at (m0.north) {\raisebox{1mm}[0mm][0mm]{Source}};
	\node[lu] at (m1.north) {\raisebox{1mm}[0mm][0mm]{Driving}};
	\node[lu] at (m2.north) {\raisebox{1mm}[0mm][0mm]{Ours}};
	\node[lu] at (m4.north) {Ours$|e|$=${\text{128}}$};
	\node[lu] at (m64.north) {Ours$|e|$=${\text{64}}$};
	\node[lu] at (m3.north) {Ours${/\text{small}\mathcal{D}}$};
	\node[lu] (label2) at (m5.north) {Ours${/2\text{-Src}}$ };	
	
	\node[lu] at  (m6.north)  {\raisebox{1mm}[0mm][0mm]{Source 2}};

	\end{tikzpicture} \vspace{-1ex}
	\caption{Ablation study in cross-reenactment on the VoxCeleb test set~\citep{voxceleb} with absolute motion transfer (upper block) and relative motion transfer (lower block). Our ablation Ours${/2\text{-Src}}$ consistently fuses the information of both source images. It produces more detail in the face, hair, and background, especially when the second source image reveals information missing in the first source image.}
	\label{fig:supp_ablation}%
\end{figure*}

\begin{figure*}
	\centering
	\begin{tikzpicture}[
	font=\sffamily\small,
	a/.style={inner sep=0.1pt},
	lu/.style={anchor=south,text=black,inner sep=2pt},
	ld/.style={anchor=north,text=black,inner sep=2pt},
	img/.style={inner sep=0pt},
	]
	\matrix (masks) [inner sep=0pt, matrix of nodes, every node/.style={a, inner sep=0pt}, column sep=0pt, row sep=0pt]
	{
		|(m0)| \includegraphics{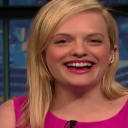}&
		|(m1)| \includegraphics{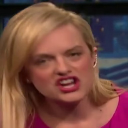}\hspace{1pt} &
		|(m2)| \includegraphics{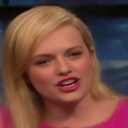}  &
		|(m3)| \includegraphics{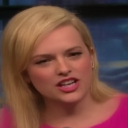} &
		|(m64)| \includegraphics{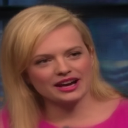} &
		|(m4)| \includegraphics{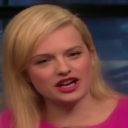} &
		|(m5)| \includegraphics{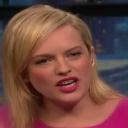}\hspace{1pt} &
		|(m6)| \includegraphics{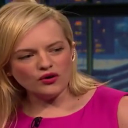} \\
		|(m7)| \includegraphics{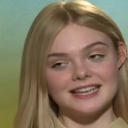} &
		|(m8)| \includegraphics{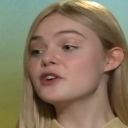}\hspace{1pt} &
		|(m9)| \includegraphics{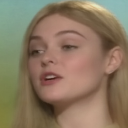} &
		|(m10)| \includegraphics{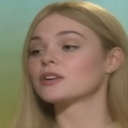} &
		|(m64_2)| \includegraphics{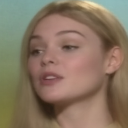} &
		|(m11)| \includegraphics{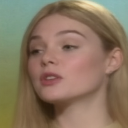} &
		|(m12)| \includegraphics{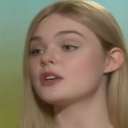}\hspace{1pt} &
		|(m13)| \includegraphics{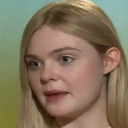} \\
		\includegraphics{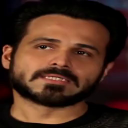} &
		\includegraphics{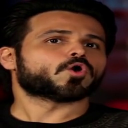}\hspace{1pt} &
		\includegraphics{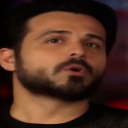} &
		\includegraphics{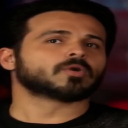} &
		\includegraphics{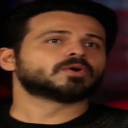} &
		\includegraphics{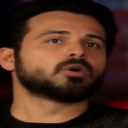} &
		\includegraphics{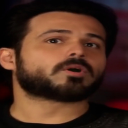}\hspace{1pt} &
		\includegraphics{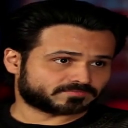} \\	
		\includegraphics{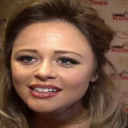} &
		\includegraphics{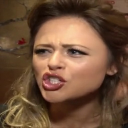}\hspace{1pt} &
		\includegraphics{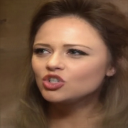} &
		\includegraphics{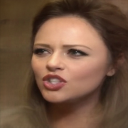} &
		\includegraphics{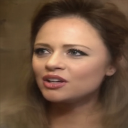} &
		\includegraphics{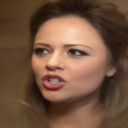} &
		\includegraphics{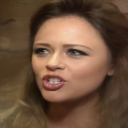}\hspace{1pt} &
		\includegraphics{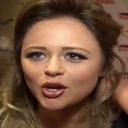} \\
		\includegraphics{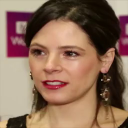} &
		\includegraphics{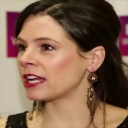}\hspace{1pt} &
		\includegraphics{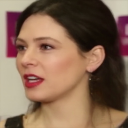} &
		\includegraphics{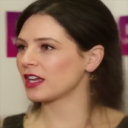} &
		\includegraphics{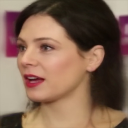} &
		\includegraphics{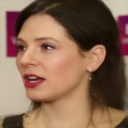} &
		\includegraphics{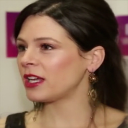}\hspace{1pt} &
		\includegraphics{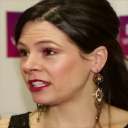} \\
		\includegraphics{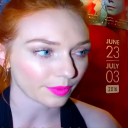} &
		\includegraphics{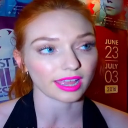}\hspace{1pt} &
		\includegraphics{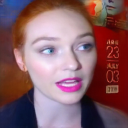} &
		\includegraphics{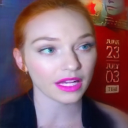} &
		\includegraphics{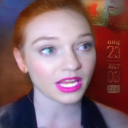} &
		\includegraphics{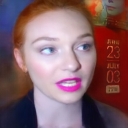} &
		\includegraphics{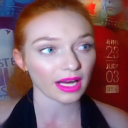}\hspace{1pt} &
		\includegraphics{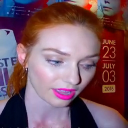} \\
		\includegraphics{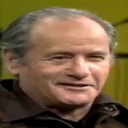} &
		\includegraphics{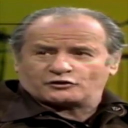}\hspace{1pt} &
		\includegraphics{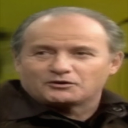} &
		\includegraphics{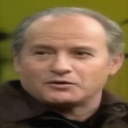} &
		\includegraphics{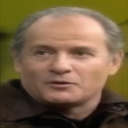} &
		\includegraphics{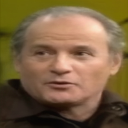} &
		\includegraphics{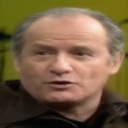}\hspace{1pt} &
		\includegraphics{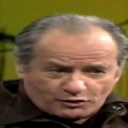} \\
		\includegraphics{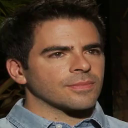} &
		\includegraphics{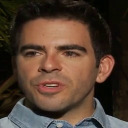}\hspace{1pt} &
		\includegraphics{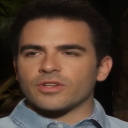} &
		\includegraphics{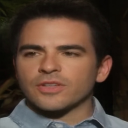} &
		\includegraphics{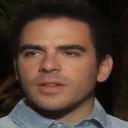} &
		\includegraphics{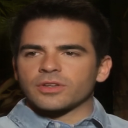} &
		\includegraphics{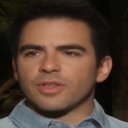}\hspace{1pt} &
		\includegraphics{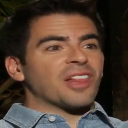} \\
		\includegraphics{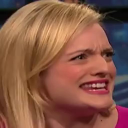} &
		\includegraphics{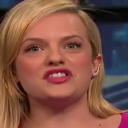}\hspace{1pt} &
		\includegraphics{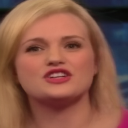} &
		\includegraphics{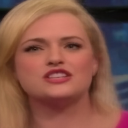} &
		\includegraphics{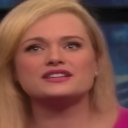} &
		\includegraphics{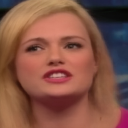} &
		\includegraphics{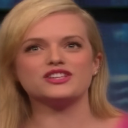}\hspace{1pt} &
		\includegraphics{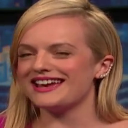} \\
	};
	
	\node[lu] at (m0.north) {\raisebox{1mm}[0mm][0mm]{Source}};
	\node[lu] at (m1.north) {\raisebox{1mm}[0mm][0mm]{Driving}};
	\node[lu] at (m2.north) {\raisebox{1mm}[0mm][0mm]{Ours}};
	\node[lu] at (m4.north) {Ours$|e|$=${\text{128}}$};
	\node[lu] at (m64.north) {Ours$|e|$=${\text{64}}$};
	\node[lu] at (m3.north) {Ours${/\text{small}\mathcal{D}}$};
	\node[lu] (label2) at (m5.north) {Ours${/2\text{-Src}}$ };	
	
	\node[lu] at (m6.north)  {\raisebox{1mm}[0mm][0mm]{Source 2}};

	\end{tikzpicture}\vspace{-1ex}
	\caption{Ablation study in self-reenactment on the VoxCeleb test set~\citep{voxceleb}. The accuracy of motion transfer (especially mouth and eye motion) decreases slightly when reducing the size of the latent expression vector $e$. In the first and fourth animation, Ours$|e|$=64 produces inaccurate mouth expressions. Ours${/2\text{-Src}}$ generates more detail by integrating the information from both source images.}
	\label{fig:supp_ablation_self}%
\end{figure*}

In \cref{fig:supp_ablation,fig:supp_ablation_self} we present qualitative results of our ablations (see \cref{sec:ablation_study}) in the cross- and self-reenactment situation, respectively.
In terms of motion transfer accuracy, our reference model with $|e|$=256 produces slightly better results than models using $|e|$=64 or $|e|$=128. 

By using two source images, information from both source images can be extracted and fused to produce more accurate animations. Especially if the second source image reveals occluded background or different head regions, less information has to be guessed by the model. As shown in \cref{fig:supp_ablation,fig:supp_ablation_self}, using multiple source images (Ours${/2\text{-Src}}$) can help to produce animations with more detail in face, hair, and background.

Our ablation with a small decoder (Ours${/\text{small}\mathcal{D}}$) has a motion transfer capability similar to our reference model (Ours), but with a slightly reduced sharpness in the animations.

\newlength{\odheight}\setlength{\odheight}{2.4cm}
\newcommand{\imgod}[1]{\includegraphics[height=\odheight,clip,trim=0 0 0 0]{#1}}
\begin{figure}
	\centering
	\begin{tikzpicture}[
	font=\sffamily\scriptsize,
	a/.style={inner sep=0.1pt},
	lu/.style={anchor=south,text=black,inner sep=0pt,minimum height=12pt},
	ld/.style={anchor=north,text=black,inner sep=0pt}
	]
	\matrix (masks) [inner sep=0pt, matrix of nodes, every node/.style={a, inner sep=0pt}, column sep=0pt, row sep=0pt]
	{
		|(m0)| \includegraphics{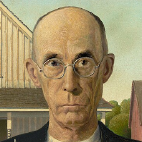}  &
		|(m1)| \includegraphics{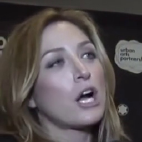}\hspace{1pt} &
		|(m2)| \includegraphics{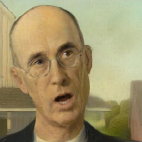} \\ [0ex]
		\includegraphics{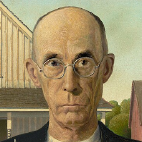}  &
		\includegraphics{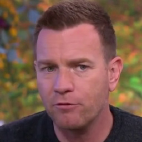}\hspace{1pt} &
		\includegraphics{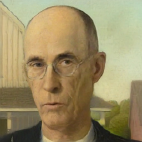} \\ [1ex]
		\includegraphics{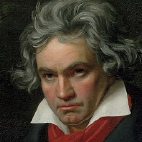} &
		\includegraphics{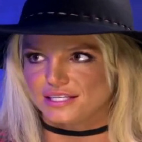}\hspace{1pt} &
		\includegraphics{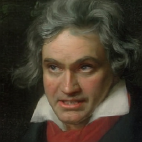} \\ [0ex]
		\includegraphics{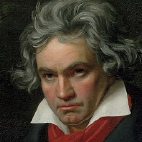} &
		\includegraphics{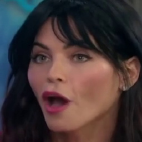}\hspace{1pt} &
		\includegraphics{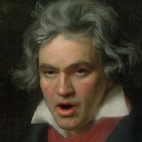} \\ [1ex]
		\includegraphics{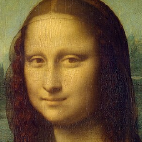} &
		\includegraphics{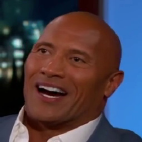}\hspace{1pt} &
		\includegraphics{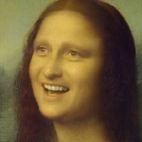} \\
		\includegraphics{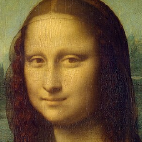} &
		\includegraphics{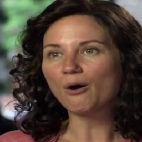}\hspace{1pt} &
		\includegraphics{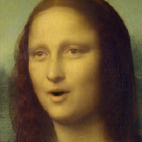} \\ 
	};

	\node[lu] at (m0.north) {\raisebox{-0.5mm}[0mm][0mm]{\small Source}};
	\node[lu] at (m1.north) {\raisebox{-0.5mm}[0mm][0mm]{\small Driving}};
	\node[lu] at (m2.north) {\raisebox{-0.5mm}[0mm][0mm]{\small Ours}};

	\end{tikzpicture} \vspace{-1ex}
	\caption{Out-of-distribution results with relative motion transfer generated by our method. The source images are extracted from popular paintings and the driving frames are from the \mbox{VoxCeleb2} test set~\cite{chung18b}.}
	\label{fig:ood}%
\end{figure}

\subsection{Comparison with State-of-the-Art Methods}
\begin{figure*}
	\centering
	\begin{tikzpicture}[
	font=\sffamily\tiny,
	a/.style={inner sep=0.1pt},
	lu/.style={anchor=base,text=black,inner sep=2pt,font=\sffamily\small},
	ld/.style={anchor=north,text=black,inner sep=2pt}
	]
	\matrix (masks) [inner sep=0pt, matrix of nodes, every node/.style={a, inner sep=0pt}, column sep=0pt, row sep=0pt]
	{
		|(m0)| \includegraphics{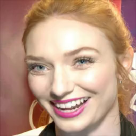}  &
		|(m1)| \includegraphics{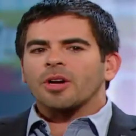}\hspace{1pt} &
		|(m2)| \includegraphics{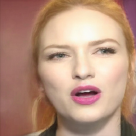}  &
		|(m3)| \includegraphics{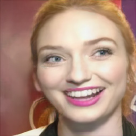} &
		|(m4)| \includegraphics{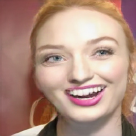} &
		|(m5)| \includegraphics{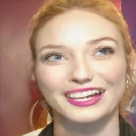} &
		|(m6)| \includegraphics{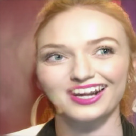} \\
		\includegraphics{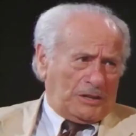} &
		\includegraphics{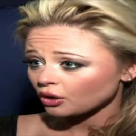}\hspace{1pt} &
		\includegraphics{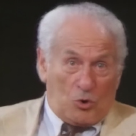} &
		\includegraphics{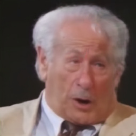} &
		\includegraphics{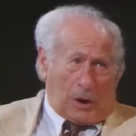} &
		\includegraphics{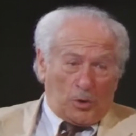} &
		\includegraphics{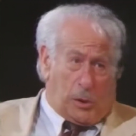} \\
		\includegraphics{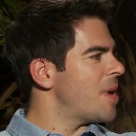} &
		\includegraphics{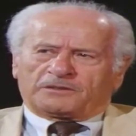}\hspace{1pt} &
		\includegraphics{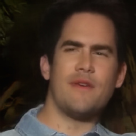} &
		\includegraphics{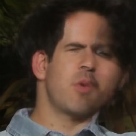} &
		\includegraphics{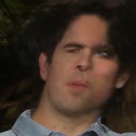} &
		\includegraphics{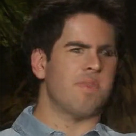} &
		\includegraphics{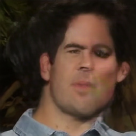}\\
		\includegraphics{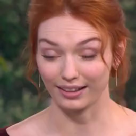} &
		\includegraphics{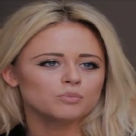}\hspace{1pt} &
		\includegraphics{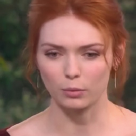} &
		\includegraphics{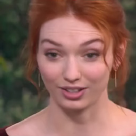} &
		\includegraphics{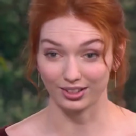} &
		\includegraphics{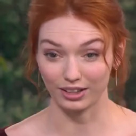} &
		\includegraphics{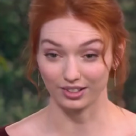}\\
		\includegraphics{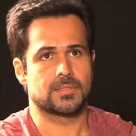} &
		\includegraphics{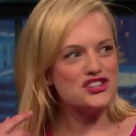}\hspace{1pt} &
		\includegraphics{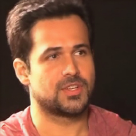} &
		\includegraphics{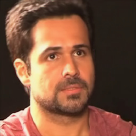} &
		\includegraphics{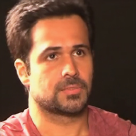} &
		\includegraphics{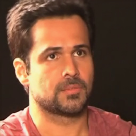} &
		\includegraphics{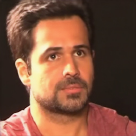}\\
		\includegraphics{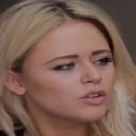} &
		\includegraphics{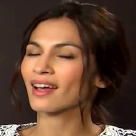}\hspace{1pt} &
		\includegraphics{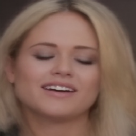} &
		\includegraphics{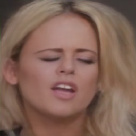} &
		\includegraphics{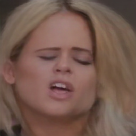} &
		\includegraphics{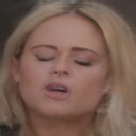} &
		\includegraphics{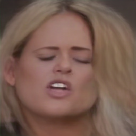}\\
		\includegraphics{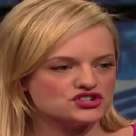} &
		\includegraphics{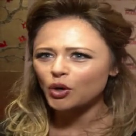}\hspace{1pt} &
		\includegraphics{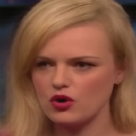} &
		\includegraphics{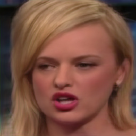} &
		\includegraphics{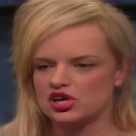} &
		\includegraphics{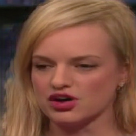} &
		\includegraphics{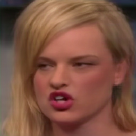}\\
		\includegraphics{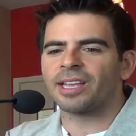} &
		\includegraphics{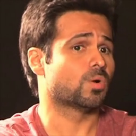}\hspace{1pt} &
		\includegraphics{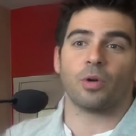} &
		\includegraphics{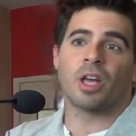} &
		\includegraphics{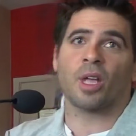} &
		\includegraphics{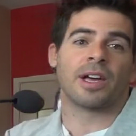} &
		\includegraphics{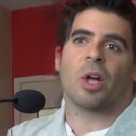}\\
		\includegraphics{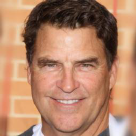} &
		\includegraphics{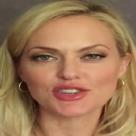}\hspace{1pt} &
	    \includegraphics{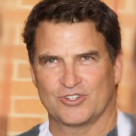} &
		\includegraphics{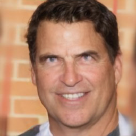} &
		\includegraphics{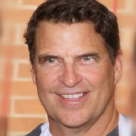} &
	    \includegraphics{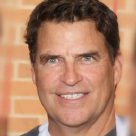} &
	    \includegraphics{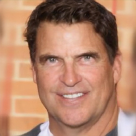}\\
	};
	
	\node[lu] at ($(m0.north)+(0,5pt)$) {Source};
	\node[lu] at ($(m1.north)+(0,5pt)$) {Driving};
	\node[lu] at ($(m2.north)+(0,5pt)$) {Ours};
	\node[lu] at ($(m3.north)+(0,5pt)$) {TSMM~\cite{spline}};
	\node[lu] at ($(m4.north)+(0,5pt)$) {DaGAN~\cite{dagan}};
	\node[lu] at ($(m5.north)+(0,5pt)$) {OSFS \cite{one-shot}};
	\node[lu] at ($(m6.north)+(0,5pt)$) {FOMM \cite{fom}};

	\end{tikzpicture}\vspace{-1ex}
	\caption{Comparison with SOTA on the VoxCeleb test set~\citep{voxceleb} in cross-reenactment (relative motion transfer).
	Our model generates more accurate expressions, is less sensitive to the alignment assumption (\cref{sec:infer}),
	and learns to realistically fill missing face parts (third row).
	Others often produce mismatched expressions and fail for large pose offsets.
	The last row shows a source image from CelebA-HQ~\cite{celebahq}.}
	\label{fig:qualitativesupprel}%
\end{figure*}

\begin{figure*}
	\centering
	\begin{tikzpicture}[
	font=\sffamily\scriptsize,
	a/.style={inner sep=0.1pt},
	lu/.style={anchor=south,text=black,inner sep=0pt,minimum height=12pt},
	ld/.style={anchor=north,text=black,inner sep=0pt}
	]
	\matrix (masks) [inner sep=0pt, matrix of nodes, every node/.style={a, inner sep=0pt}, column sep=0pt, row sep=0pt]
	{
		|(m0)| \includegraphics{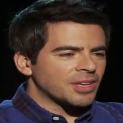}  &
		|(m1)| \includegraphics{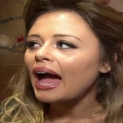}\hspace{1pt} &
		|(m2)| \includegraphics{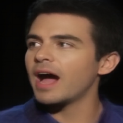}  &
		|(m3)| \includegraphics{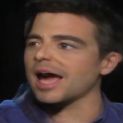} &
		|(m4)| \includegraphics{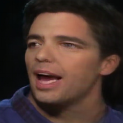} &
		|(m5)| \includegraphics{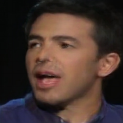} &
		|(m6)| \includegraphics{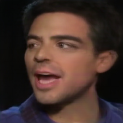} &
		|(m7)| \includegraphics{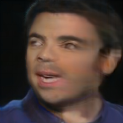} \\
		\includegraphics{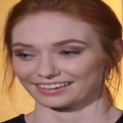} &
		\includegraphics{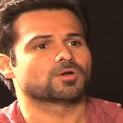}\hspace{1pt} &
		\includegraphics{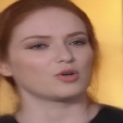} &
		\includegraphics{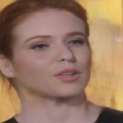} &
		\includegraphics{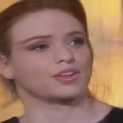} &
		\includegraphics{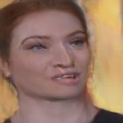} &
		\includegraphics{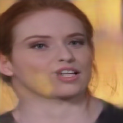} &
		\includegraphics{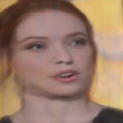}\\
		\includegraphics{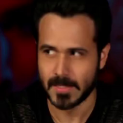} &
		\includegraphics{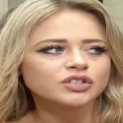}\hspace{1pt} &
		\includegraphics{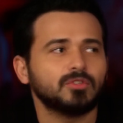} &
		\includegraphics{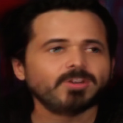} &
		\includegraphics{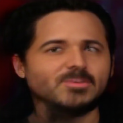} &
		\includegraphics{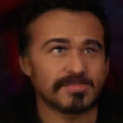} &
		\includegraphics{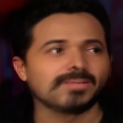} &
		\includegraphics{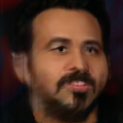} \\
		\includegraphics{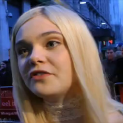} &
		\includegraphics{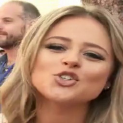}\hspace{1pt} &
		\includegraphics{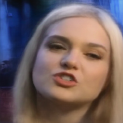} &
		\includegraphics{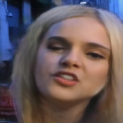} &
		\includegraphics{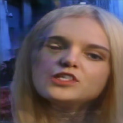} &
		\includegraphics{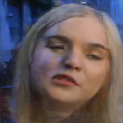} &
		\includegraphics{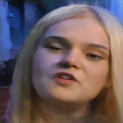} &
		\includegraphics{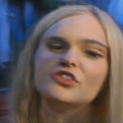}\\
		\includegraphics{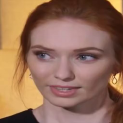} &
		\includegraphics{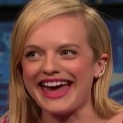}\hspace{1pt} &
		\includegraphics{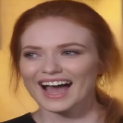} &
		\includegraphics{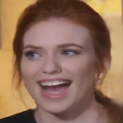} &
		\includegraphics{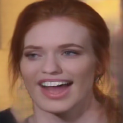} &
		\includegraphics{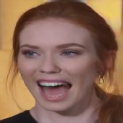} &
		\includegraphics{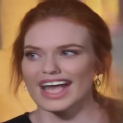} &
		\includegraphics{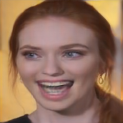}\\
		\includegraphics{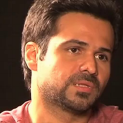} &
		\includegraphics{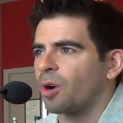}\hspace{1pt} &
		\includegraphics{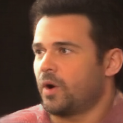} &
		\includegraphics{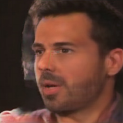} &
		\includegraphics{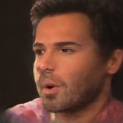} &
		\includegraphics{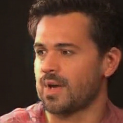} &
		\includegraphics{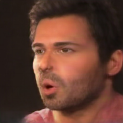} &
		\includegraphics{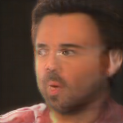}\\
		\includegraphics{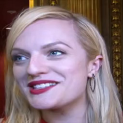} &
		\includegraphics{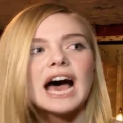}\hspace{1pt} &
		\includegraphics{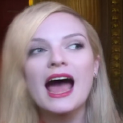} &
		\includegraphics{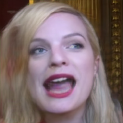} &
		\includegraphics{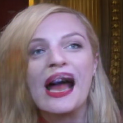} &
		\includegraphics{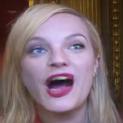} &
		\includegraphics{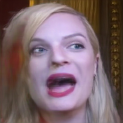} &
		\includegraphics{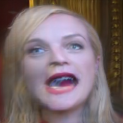}\\
		\includegraphics{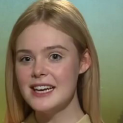} &
		\includegraphics{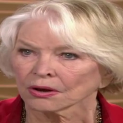}\hspace{1pt} &
		\includegraphics{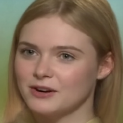} &
		\includegraphics{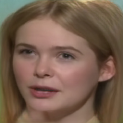} &
		\includegraphics{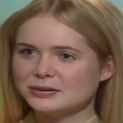} &
		\includegraphics{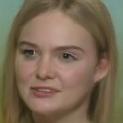} &
		\includegraphics{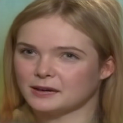} &
		\includegraphics{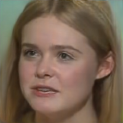}\\
		\includegraphics{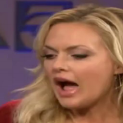} &
		\includegraphics{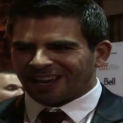}\hspace{1pt} &
		\includegraphics{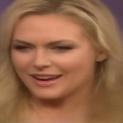} &
		\includegraphics{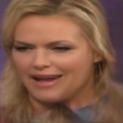} &
		\includegraphics{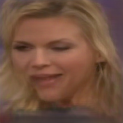} &
		\includegraphics{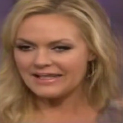} &
		\includegraphics{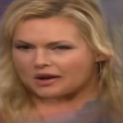} &
		\includegraphics{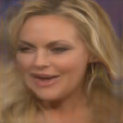}\\
		\includegraphics{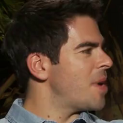} &
		\includegraphics{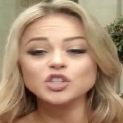}\hspace{1pt} &
		\includegraphics{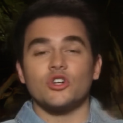} &
		\includegraphics{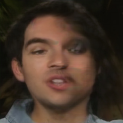} &
		\includegraphics{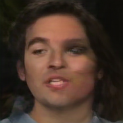} &
		\includegraphics{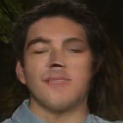} &
		\includegraphics{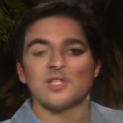} &
		\includegraphics{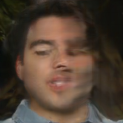}\\
	};
	
	\node[lu] at (m0.north) {\raisebox{-0.5mm}[0mm][0mm]{\small Source}};
	\node[lu] at (m1.north) {\raisebox{-0.5mm}[0mm][0mm]{\small Driving}};
	\node[lu] at (m2.north) {\raisebox{-0.5mm}[0mm][0mm]{\small Ours}};
	\node[lu] at (m3.north) {\small TSMM~\cite{spline}};
	\node[lu] at (m4.north) {\small DaGAN~\cite{dagan}};
	\node[lu] at (m5.north) {\small OSFS \cite{one-shot}};
	\node[lu] at (m6.north) {\small FOMM \cite{fom}};
	\node[lu] at (m7.north) {\small DPE \cite{dpe}};

	\end{tikzpicture} \vspace{-1.5ex}
	\caption{Comparison with SOTA on the VoxCeleb test set~\citep{voxceleb} in cross-reenactment with absolute motion transfer. We generate more accurate facial expressions with better ID preservation. Related methods often produce strong shape deformations, artifacts and blurry results (especially in the mouth region). The sixth animation shows that our method even animates the sunlight on the side of the face.}
	\label{fig:qualitativesuppabs}%
\end{figure*}

In \cref{fig:qualitativesupprel} and \cref{fig:qualitativesuppabs} we present additional cross-reenactment results on the VoxCeleb test set~\citep{voxceleb} with relative and absolute motion transfer compared to all state-of-the-art methods from our user study (see \cref{tab:study}). While TSMM~\cite{spline},
DaGAN~\cite{dagan}, OSFS~\cite{one-shot}, and FOMM~\cite{fom} are also keypoint based, DPE~\cite{dpe} uses a latent head pose description. This, however, eliminates the ability to perform relative motion transfer.
As the visualizations show, our method produces significantly more natural results with higher ID preservation and more accurate and plausible motion transfers. Especially when there is a large pose offset, related methods often fail to produce satisfactory results.
For animated results, see our project page.\footnote{\url{https://andrerochow.github.io/fsrt}}

\subsection{Out-of-Distribution Animation}
As shown in \cref{fig:ood}, our model trained on VoxCeleb~\cite{voxceleb} generalizes to out-of-distribution source images extracted from popular paintings. 

\subsection{Generalizing to other Datasets}
We report generalization examples of our models trained on VoxCeleb to other datasets at inference time.
Specifically, we show the following source $\rightarrow$ driving combinations:
{\setlength\tabcolsep{2pt}
\begin{tabular}{@{\hspace{4pt}}llcl}
  $\bullet$ & CelebA-HQ~\cite{celebahq}          & $\rightarrow$ & VoxCeleb2~\cite{chung18b} in \cref{fig:vox2CelebA}, \\
  $\bullet$ & VoxCeleb2~\cite{chung18b}          & $\rightarrow$ & VoxCeleb2~\cite{chung18b} in \cref{fig:vox2}, and \\
  $\bullet$ & CelebV~\cite{wu2018reenactgan}     & $\rightarrow$ & CelebV~\cite{wu2018reenactgan} in \cref{fig:celebV}. \\
\end{tabular}}\\
We note that VoxCeleb2 covers a significantly larger number of identities in the test set compared to VoxCeleb.
As the results show, our model generalizes to all of these combinations, while still producing more accurate animations compared to related methods.

\subsection{Omitting Keypoints}
\begin{figure*}
	\centering
	\begin{tikzpicture}[
	font=\sffamily\scriptsize,
	a/.style={inner sep=0.1pt},
	lu/.style={anchor=south,text=black,inner sep=0pt,minimum height=12pt},
	ld/.style={anchor=north,text=black,inner sep=0pt}
	]
	\matrix (masks) [inner sep=0pt, matrix of nodes, every node/.style={a, inner sep=0pt}, column sep=0pt, row sep=0pt]
	{
		|(m0)| \includegraphics{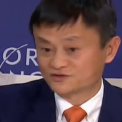}  &
		|(m1)| \includegraphics{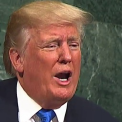}\hspace{1pt} &
		|(m2)| \includegraphics{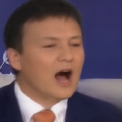}  &
		|(m6)| \includegraphics{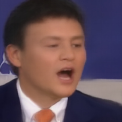}  &
		|(m3)| \includegraphics{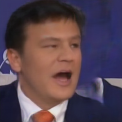} &
		|(m4)| \includegraphics{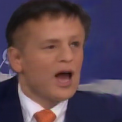} &
		|(m5)| \includegraphics{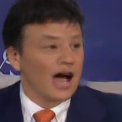} &
		|(m7)| \includegraphics{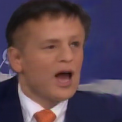} \\
		\includegraphics{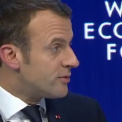} &
		\includegraphics{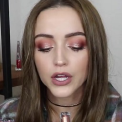}\hspace{1pt} &
		\includegraphics{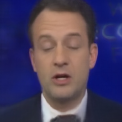} &
		\includegraphics{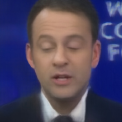} &
		\includegraphics{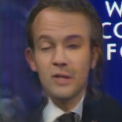} &
		\includegraphics{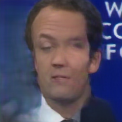} &
		\includegraphics{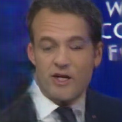} &
		\includegraphics{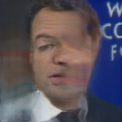} \\
		\includegraphics{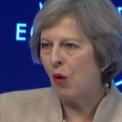} &
		\includegraphics{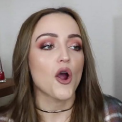}\hspace{1pt} &
		\includegraphics{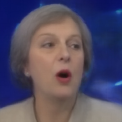} &
		\includegraphics{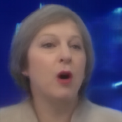} &
		\includegraphics{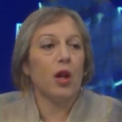} &
		\includegraphics{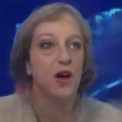} &
		\includegraphics{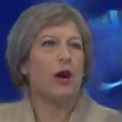} &
		\includegraphics{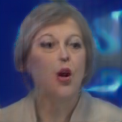}\\
		\includegraphics{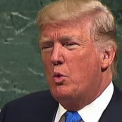} &
		\includegraphics{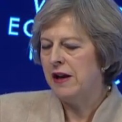}\hspace{1pt} &
		\includegraphics{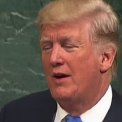} &
		\includegraphics{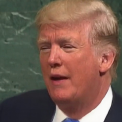} &
		\includegraphics{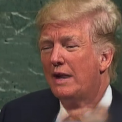} &
		\includegraphics{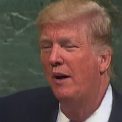} &
		\includegraphics{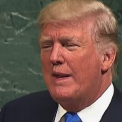} &
		\includegraphics{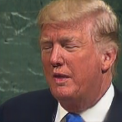}\\
		|(mb0)| \includegraphics{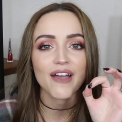} &
		\includegraphics{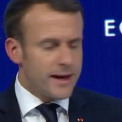}\hspace{1pt} &
		\includegraphics{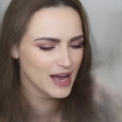} &
		\includegraphics{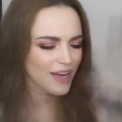} &
		\includegraphics{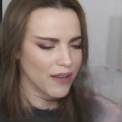} &
		\includegraphics{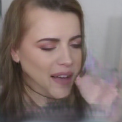} &
		\includegraphics{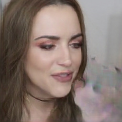} &
		\includegraphics{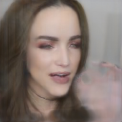}\\[0.2cm]
		|(b0)| \includegraphics{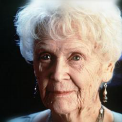} &
		|(b1)| \includegraphics{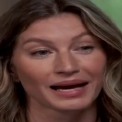}\hspace{1pt} &
		|(b2)| \includegraphics{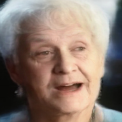} &
		|(b6)| \includegraphics{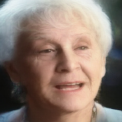} &
		|(b3)| \includegraphics{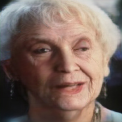} &
		|(b4)| \includegraphics{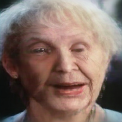} &
		|(b5)| \includegraphics{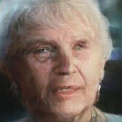} &
		|(b7)| \includegraphics{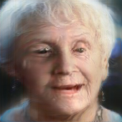}\\
		\includegraphics{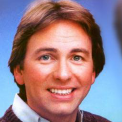} &
		\includegraphics{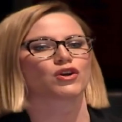}\hspace{1pt} &
		\includegraphics{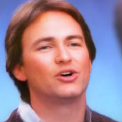} &
		\includegraphics{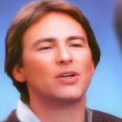} &
		\includegraphics{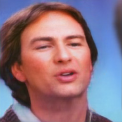} &
		\includegraphics{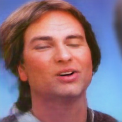} &
		\includegraphics{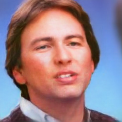} &
		\includegraphics{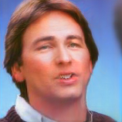}\\
		\includegraphics{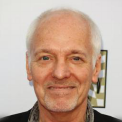} &
		\includegraphics{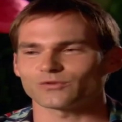}\hspace{1pt} &
		\includegraphics{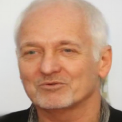} &
		\includegraphics{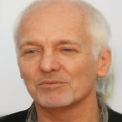} &
		\includegraphics{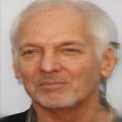} &
		\includegraphics{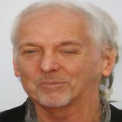} &
		\includegraphics{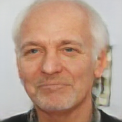} &
		\includegraphics{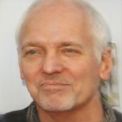}\\
		\includegraphics{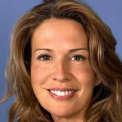} &
		\includegraphics{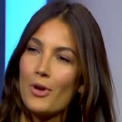}\hspace{1pt} &
		\includegraphics{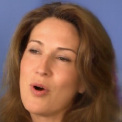} &
		\includegraphics{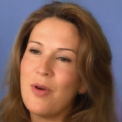} &
		\includegraphics{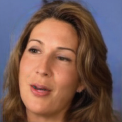} &
		\includegraphics{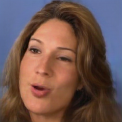} &
		\includegraphics{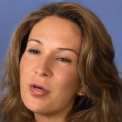} &
		\includegraphics{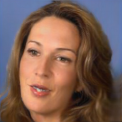}\\
		|(bb0)| \includegraphics{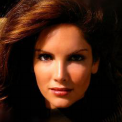} &
		\includegraphics{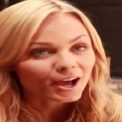}\hspace{1pt} &
		\includegraphics{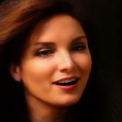} &
		\includegraphics{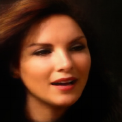} &
		\includegraphics{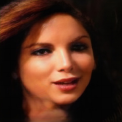} &
		\includegraphics{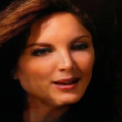} &
		\includegraphics{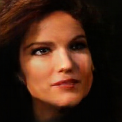} &
		\includegraphics{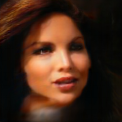}\\
	};
	
	\node[lu] at (m0.north) {\raisebox{-0.5mm}[0mm][0mm]{\small Source}};
	\node[lu] at (m1.north) {\raisebox{-0.5mm}[0mm][0mm]{\small Driving}};
	\node[lu] at (m2.north) {\raisebox{-0.5mm}[0mm][0mm]{\small Ours}};
    \node[lu] at (m6.north) {Ours$/$\small $n_\mathcal{K}=0$};
	\node[lu] at (m3.north) {\small TSMM~\cite{spline}};
	\node[lu] at (m4.north) {\small DaGAN~\cite{dagan}};
	\node[lu] at (m5.north) {\small OSFS \cite{one-shot}};

	\node[lu] at (m7.north) {\small DPE \cite{dpe}};

	\coordinate (topblock) at ($(m0.west)!0.5!(mb0.west)$);
	\node [anchor=base,rotate=90,font=\sffamily\small] at ($(topblock)+(-5pt,0)$) {CelebV~\cite{wu2018reenactgan} $\rightarrow$ CelebV~\cite{wu2018reenactgan}};

	\coordinate (botblock) at ($(b0.west)!0.5!(bb0.west)$);
	\node [anchor=base,rotate=90,font=\sffamily\small] at ($(botblock)+(-5pt,0)$) {CelebA-HQ~\citep{celebahq} $\rightarrow$ VoxCeleb2~\cite{chung18b}};

	\end{tikzpicture} \vspace{-1.5ex}
	\caption{Comparison of our model with and without keypoints and state-of-the-art methods in cross-reenactment with absolute motion transfer.
	The top block shows generalization to source and driving frames extracted from the CelebV dataset~\cite{wu2018reenactgan}.
	The bottom block shows generalization to driving frames extracted from the VoxCeleb2 test set~\cite{chung18b} and source images from the CelebA-HQ dataset~\citep{celebahq}.}
	\label{fig:celebV}%
\end{figure*}

We present qualitative results of our model ablation Ours$/$$n_\mathcal{K}=0$ without keypoints in \cref{fig:celebV}. Compared to our reference model (Ours), we found that the accuracy of the motion transfer is slightly reduced. In particular, the animated gaze direction seems to be less accurate (see third row in \cref{fig:celebV}). Omitting the keypoints makes it impossible to perform relative motion transfer, since all pose information is implicitly encoded in the expression vector $e$.

\begin{figure}[t]
	\centering
	\begin{tikzpicture}[
	font=\sffamily\scriptsize,
	a/.style={inner sep=0.1pt},
	lu/.style={anchor=base,text=black,inner sep=0pt,minimum height=12pt},
	ld/.style={anchor=north,text=black,inner sep=0pt}
	]
	\matrix (masks) [inner sep=0pt, matrix of nodes, every node/.style={a, inner sep=0pt}, column sep=0pt, row sep=0pt]
	{
		|(m0)| \includegraphics{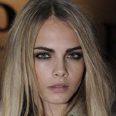}  &
		|(m1)| \includegraphics{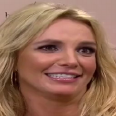}\hspace{1pt} &
		|(m2)| \includegraphics{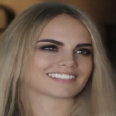} & 
		|(m3)| \includegraphics{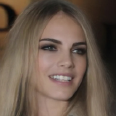}\\ 
		|(sv2_0)|\includegraphics{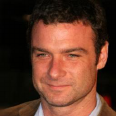} &
		\includegraphics{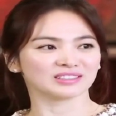}\hspace{1pt} &
		\includegraphics{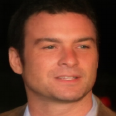} &
		\includegraphics{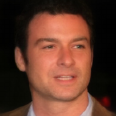} \\
		|(sv2_1)|\includegraphics{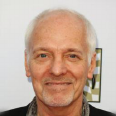} &
		\includegraphics{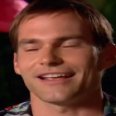}\hspace{1pt} &
		\includegraphics{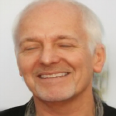} &
		\includegraphics{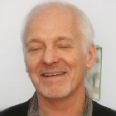} \\
		\includegraphics{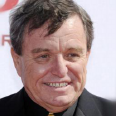} &
		\includegraphics{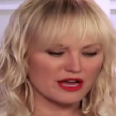}\hspace{1pt} &
		\includegraphics{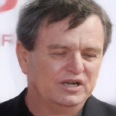} &
		\includegraphics{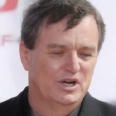} \\ [1ex]
		|(sv1_0)|\includegraphics{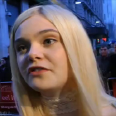} &
		\includegraphics{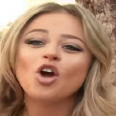}\hspace{1pt} &
		\includegraphics{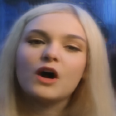} &
		\includegraphics{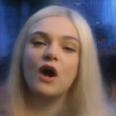} \\
		|(sv1_1)|\includegraphics{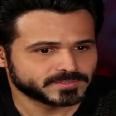} &
		\includegraphics{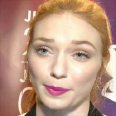}\hspace{1pt} &
		\includegraphics{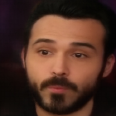} &
		\includegraphics{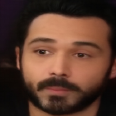} \\
	};

	\coordinate (topblock) at ($(sv2_0.west)!0.5!(sv2_1.west)$);
\node [anchor=base,rotate=90,font=\sffamily\scriptsize] at ($(topblock)+(-3pt,0)$) {CelebA-HQ~\citep{celebahq} $\rightarrow$ VoxCeleb2~\cite{chung18b}};

	\coordinate (topblock) at ($(sv1_0.west)!0.5!(sv1_1.west)$);
\node [anchor=base,rotate=90,font=\sffamily\scriptsize] at ($(topblock)+(-3pt,0)$) {VoxCeleb~\cite{voxceleb} $\rightarrow$ VoxCeleb~\cite{voxceleb}};
	\begin{scope}[font=\sffamily\footnotesize]
	\node[lu] at ($(m0.north)+(0,4pt)$) {Source};
	\node[lu] at ($(m1.north)+(0,4pt)$) {Driving};
	\node[lu] at ($(m2.north)+(0,4pt)$) {Ours$/$$n_\mathcal{K}=0$};
	\node[lu] at ($(m3.north)+(0,4pt)$) {\adjustbox{raise=0.25ex}{+}\,Crop Aug.};
	\end{scope}

	\end{tikzpicture}
	\caption{Ablations without keypoints. This comparison is using absolute motion transfer.
	When combining a keypoint-less model with random center cropping during training (right column), shape deformations and scale changes are prevented.
	The images are from the VoxCeleb test set~\cite{voxceleb}, the VoxCeleb2 test set~\cite{chung18b}, and the CelebA-HQ dataset~\citep{celebahq} (as indicated by the source $\rightarrow$ driving notation).
	}
	\label{fig:center_crop}%
	\vspace{1.4cm}
\end{figure}

In this variant, images input to the expression network are not augmented through cropping, since this makes recovery of the head pose impossible without keypoints. However, we discovered that performing a random center crop with variable aspect ratio on the driving frame (while requiring the network to reconstruct the full driving frame) reduces shape deformations, since the network becomes invariant against aspect ratio changes and scale (see Ours$/$$n_{\mathcal{K}}=0$ + Crop Aug. in \cref{fig:center_crop}).
While this might be useful in cross-reenactment applications where relative motion transfer is not required,
it reduces self-reenactment scores (see \cref{tab:noKP})---where this invariance is not helpful but actually harmful.
A particular reason for this might be that this variant cannot transfer zooming or dolly shots due to scale invariance.

\subsection{Statistical Regularization}
\begin{figure*}\sf \tiny \centering
	\begin{tikzpicture}
	  \matrix[matrix of nodes, inner sep=0cm] (m) {
	    Source 1 & Source 2 & Source 3 & Source 4 \\[.1cm]
	    \includegraphics{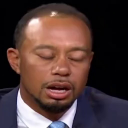} &
	    \includegraphics{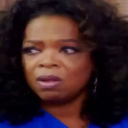} &
	    \includegraphics{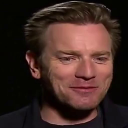} &
	    \includegraphics{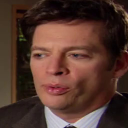} \\[0.1cm]
	    |(pred1)| \includegraphics{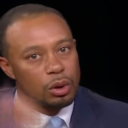} &
	    \includegraphics{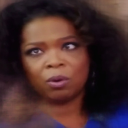} &
	    \includegraphics{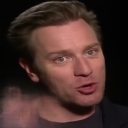} &
	    \includegraphics{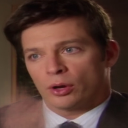} \\
	    |(pred2)| \includegraphics{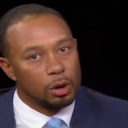} &
	    \includegraphics{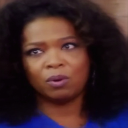} &
	    \includegraphics{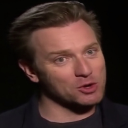} &
	    \includegraphics{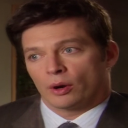} \\
	  };
	  \coordinate (midpred) at ($(pred1.west)!0.5!(pred2.west)$);
	\node[left=1cm of midpred,label={below:Driving frame}] {\includegraphics{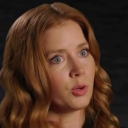}};

	  \node[rotate=90,anchor=south] at (pred1.west) {w$/$o Stat. Reg.};
	  \node[rotate=90,anchor=south] at (pred2.west) {Ours};

	  \draw[latex-,very thick,red] (rel cs:x=32,y=10,name=pred1) -- ++(0.5cm, 0);
	  \draw[latex-,very thick,red] (rel cs:x=30,y=10,name=m-3-2) -- ++(0.5cm, 0);
	  \draw[latex-,very thick,red] (rel cs:x=32,y=10,name=m-3-3) -- ++(0.5cm, 0);
	  \draw[latex-,very thick,red] (rel cs:x=33,y=10,name=m-3-4) -- ++(0.5cm, 0);
	\end{tikzpicture}

\vspace{0.25cm}

	\begin{tikzpicture}
\matrix[matrix of nodes, inner sep=0cm] {
	Source 1 & Source 2* & Source 3 & Source 4 \\[.1cm]
	\includegraphics{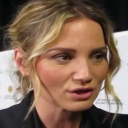} &
	\includegraphics{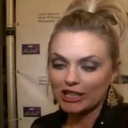} &
	\includegraphics{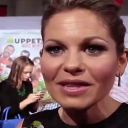} &
	\includegraphics{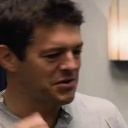} \\[0.1cm]
	|(pred1)| \includegraphics{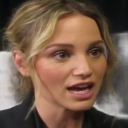} &
	\includegraphics{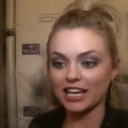} &
	\includegraphics{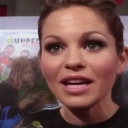} &
	\includegraphics{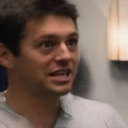} \\
	|(pred2)| \includegraphics{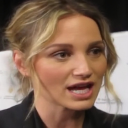} &
	\includegraphics{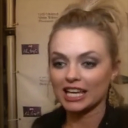} &
	\includegraphics{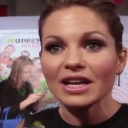} &
	\includegraphics{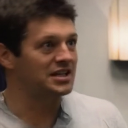} \\
};
\coordinate (midpred) at ($(pred1.west)!0.5!(pred2.west)$);
\node[left=1cm of midpred,label={below:Driving frame*}] {\includegraphics{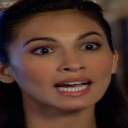}};

\node[rotate=90,anchor=south] at (pred1.west) {w$/$o Stat. Reg.};
\node[rotate=90,anchor=south] at (pred2.west) {Ours};

\draw[latex-,very thick,red] (rel cs:x=89,y=23,name=pred1) -- ++(249:0.5cm);
\draw[latex-,very thick,red] (rel cs:x=27,y=42,name=m-3-2) -- ++(120:0.5cm);
\draw[latex-,very thick,red] (rel cs:x=26,y=31,name=m-3-3) -- ++(208:0.5cm);
\draw[latex-,very thick,red] (rel cs:x=93,y=20,name=m-3-3) -- ++(0, -0.4cm);
\draw[latex-,very thick,red] (rel cs:x=71,y=40,name=m-3-4) -- ++(0.5cm, 0);
\end{tikzpicture}

\vspace{0.25cm}

	\begin{tikzpicture}
\matrix[matrix of nodes, inner sep=0cm] {
	Source 1 & Source 2* & Source 3 & Source 4 \\[.1cm]
	\includegraphics{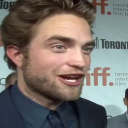} &
	\includegraphics{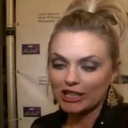} &
	\includegraphics{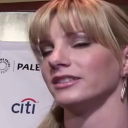} &
	\includegraphics{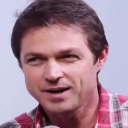} \\[0.1cm]
	|(pred1)| \includegraphics{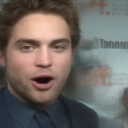} &
	\includegraphics{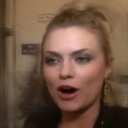} &
	\includegraphics{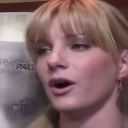} &
	\includegraphics{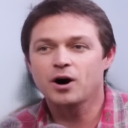} \\
	|(pred2)| \includegraphics{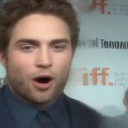} &
	\includegraphics{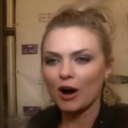} &
	\includegraphics{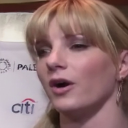} &
	\includegraphics{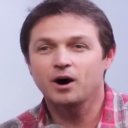} \\
};
\coordinate (midpred) at ($(pred1.west)!0.5!(pred2.west)$);
\node[left=1cm of midpred,label={below:Driving frame}] {\includegraphics{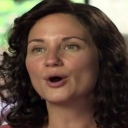}};

\node[rotate=90,anchor=south] at (pred1.west) {w$/$o Stat. Reg.};
\node[rotate=90,anchor=south] at (pred2.west) {Ours};

\draw[latex-,very thick,red] (rel cs:x=68,y=40,name=pred1) -- ++(-10:0.5cm);
\draw[latex-,very thick,red] (rel cs:x=20,y=40,name=m-3-2) -- ++(120:0.5cm);
\draw[latex-,very thick,red] (rel cs:x=23,y=30,name=m-3-3) -- ++(160:0.5cm);
\draw[latex-,very thick,red] (rel cs:x=17,y=40,name=m-3-4) -- ++(130:0.5cm);
\end{tikzpicture}

	\vspace{-1.5ex}
	\caption{Benefit of statistical regularization (relative motion transfer). Training without $\mathcal{L}_\text{Cov}$ and $\mathcal{L}_\text{Var}$ leads to visible artifacts around the animated face (see red arrows), indicating that the identity of the driving person is leaking into the expression vector $e_D$. The images are from the VoxCeleb test set~\cite{voxceleb} (indicated with *) and the VoxCeleb2 test set~\cite{chung18b} (remaining).}

	\label{fig:noreg}
\end{figure*}

In \cref{fig:noreg}, we visualize the effect of training without our proposed statistical regularization. As the results show, training without $\mathcal{L}_{\text{Cov}}$ and $\mathcal{L}_{\text{Var}}$ leads to significant artifacts around the animated face region, indicating that ID information leaks from the driving frame through the expression vector $e_D$. Our proposed factorization is therefore not achieved.

\makeatletter
\setlength{\@fptop}{0pt}
\makeatother

\begin {table} \centering \footnotesize  %
\begin{ThreePartTable}
	\begin{tabu} to \linewidth {lr@{\hspace{18pt}}rrrr}
		\toprule

		Method                              & \#KP & SSIM$\uparrow$    & PSNR$\uparrow$     & L1$\downarrow$   & AKD$\downarrow$ \\

		\midrule
		Ours                                & 10 & .7576             & 23.67              & .0421 & 2.13  \\
		\hspace{0.5em} $n_{\mathcal{K}}=0$  & 0  & .7445             & 23.56              & .0436            & 2.64  \\
		\hspace{1em} +Crop Aug.             & 0  & .7240             & 22.98              & .0469            & 2.99  \\
		\bottomrule
	\end{tabu}
	\caption{Self-reenactment results on the official VoxCeleb test set~\cite{voxceleb}. We compare our model ablation without keypoints (Ours$/$$n_{\mathcal{K}}=0$) with an ablation that is additionally trained with random center cropping (Ours$/$$n_{\mathcal{K}}=0$ + Crop Aug.). The scores of our reference model (Ours) are shown in the first row.}
	\label{tab:noKP}
\end{ThreePartTable}
\end{table}

\newlength{\nkheight}\setlength{\nkheight}{1.95cm}	
\newcommand{\imgnk}[1]{\includegraphics[height=\nkheight,clip,trim=0 0 0 0]{#1}}
	
\newlength{\sh}\setlength{\sh}{2.11cm}
\newlength{\sfheight}\setlength{\sfheight}{2.155cm}	
\newcommand{\imgsf}[1]{\includegraphics[height=\sfheight,clip,trim=0 0 0 0]{#1}}

\newlength{\sfnheight}\setlength{\sfnheight}{2.155cm}	
\newcommand{\imgsfn}[1]{\includegraphics[height=\sfnheight,clip,trim=0 0 0 0]{#1}}

\newlength{\cidsheight}\setlength{\cidsheight}{2.3cm}
\newcommand{\imgcids}[1]{\includegraphics[height=\cidsheight,clip,trim=0 0 0 0]{#1}}

\newlength{\cidsaheight}\setlength{\cidsaheight}{2.081cm}
\newcommand{\imgcidsa}[1]{\includegraphics[height=\cidsaheight,clip,trim=0 0 0 0]{#1}}

\begin{figure*}
	\centering
	\begin{tikzpicture}[
	font=\sffamily\tiny,
	a/.style={inner sep=0.1pt},
	lu/.style={anchor=south,text=black,inner sep=2pt},
	ld/.style={anchor=north,text=black,inner sep=2pt}
	]
	\matrix (masks) [inner sep=0pt, matrix of nodes, every node/.style={a, inner sep=0pt}, column sep=0pt, row sep=0pt]
	{
		|(m0)| \includegraphics{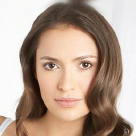}  &
		|(m1)| \includegraphics{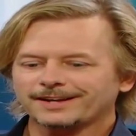}\hspace{1pt} &
		|(m2)| \includegraphics{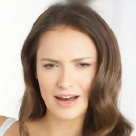}  &
		|(m3)| \includegraphics{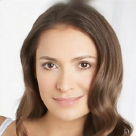} &
		|(m4)| \includegraphics{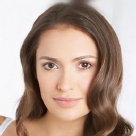} &
		|(m5)| \includegraphics{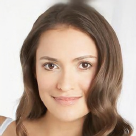} &
		|(m6)| \includegraphics{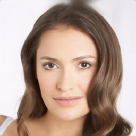} \\
		\includegraphics{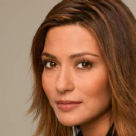} &
		\includegraphics{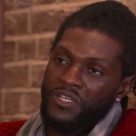}\hspace{1pt} &
		\includegraphics{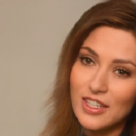} &
		\includegraphics{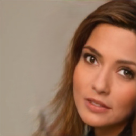} &
		\includegraphics{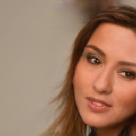} &
		\includegraphics{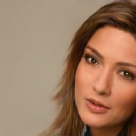} &
		\includegraphics{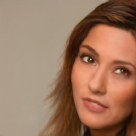}\\
		\includegraphics{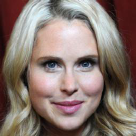} &
		\includegraphics{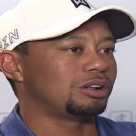}\hspace{1pt} &
		\includegraphics{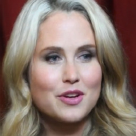} &
		\includegraphics{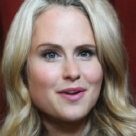} &
		\includegraphics{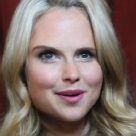} &
		\includegraphics{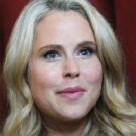} &
		\includegraphics{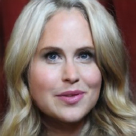} \\
		\includegraphics{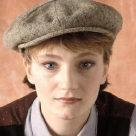} &
		\includegraphics{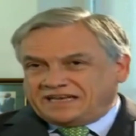}\hspace{1pt} &
		\includegraphics{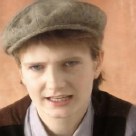} &
		\includegraphics{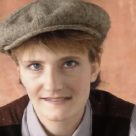} &
		\includegraphics{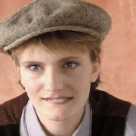} &
		\includegraphics{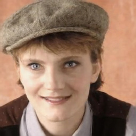} &
		\includegraphics{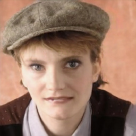}\\
		\includegraphics{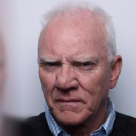} &
		\includegraphics{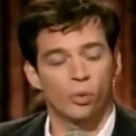}\hspace{1pt} &
		\includegraphics{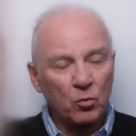} &
		\includegraphics{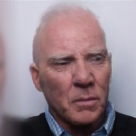} &
		\includegraphics{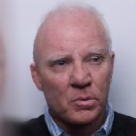} &
		\includegraphics{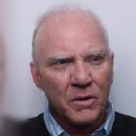} &
		\includegraphics{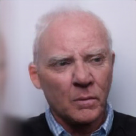}\\
		\includegraphics{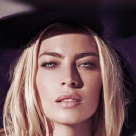} &
		\includegraphics{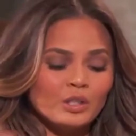}\hspace{1pt} &
		\includegraphics{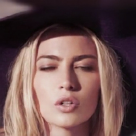} &
		\includegraphics{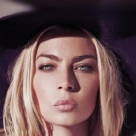} &
		\includegraphics{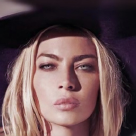} &
		\includegraphics{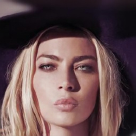} &
		\includegraphics{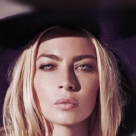}\\
		\includegraphics{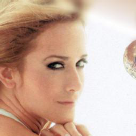} &
		\includegraphics{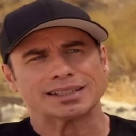}\hspace{1pt} &
		\includegraphics{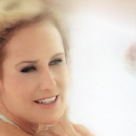} &
		\includegraphics{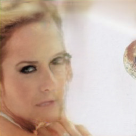} &
		\includegraphics{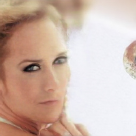} &
		\includegraphics{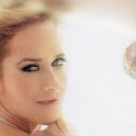} &
		\includegraphics{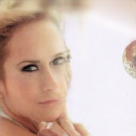}\\
		\includegraphics{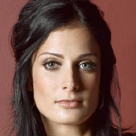} &
		\includegraphics{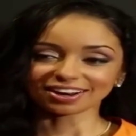}\hspace{1pt} &
		\includegraphics{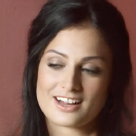} &
		\includegraphics{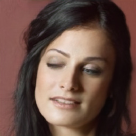} &
		\includegraphics{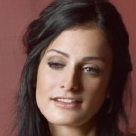} &
		\includegraphics{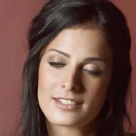} &
		\includegraphics{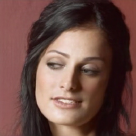}\\
		\includegraphics{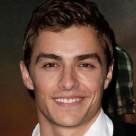} &
		\includegraphics{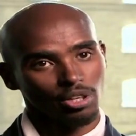}\hspace{1pt} &
		\includegraphics{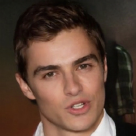} &
		\includegraphics{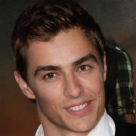} &
		\includegraphics{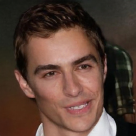} &
		\includegraphics{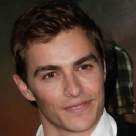} &
		\includegraphics{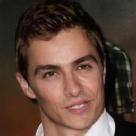}\\
	};
	
	\node[lu] at (m0.north) {\raisebox{1mm}[0mm][0mm]{\small Source}};
	\node[lu] at (m1.north) {\raisebox{1mm}[0mm][0mm]{\small Driving}};
	\node[lu] at (m2.north) {\raisebox{1mm}[0mm][0mm]{\small Ours}};
	\node[lu] at (m3.north) {\small TSMM~\cite{spline}};
	\node[lu] at (m4.north) {\small DaGAN~\cite{dagan}};
	\node[lu] at (m5.north) {\small OSFS \cite{one-shot}};
	\node[lu] at (m6.north) {\small FOMM \cite{fom}};

	\end{tikzpicture}\vspace{-1ex}
	\caption{ %
	Cross-reenactment generalization to driving videos from the VoxCeleb2 test set~\cite{chung18b} and source images from the CelebA-HQ dataset~\cite{celebahq} with relative motion transfer.
	}
	\label{fig:vox2CelebA}%
\end{figure*}

\begin{figure*}
	\centering
	\begin{tikzpicture}[
	font=\sffamily\tiny,
	a/.style={inner sep=0.1pt},
	lu/.style={anchor=south,text=black,inner sep=2pt},
	ld/.style={anchor=north,text=black,inner sep=2pt}
	]
	\matrix (masks) [inner sep=0pt, matrix of nodes, every node/.style={a, inner sep=0pt}, column sep=0pt, row sep=0pt]
	{
		|(m0)| \includegraphics{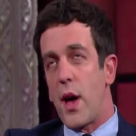}  &
		|(m1)| \includegraphics{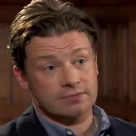}\hspace{1pt} &
		|(m2)| \includegraphics{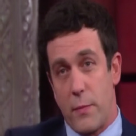}  &
		|(m3)| \includegraphics{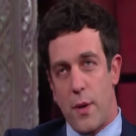} &
		|(m4)| \includegraphics{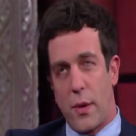} &
		|(m5)| \includegraphics{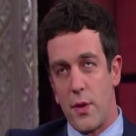} &
		|(m6)| \includegraphics{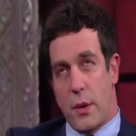} \\
		\includegraphics{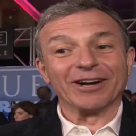} &
		\includegraphics{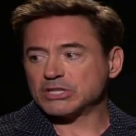}\hspace{1pt} &
		\includegraphics{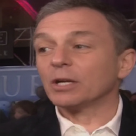} &
		\includegraphics{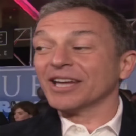} &
		\includegraphics{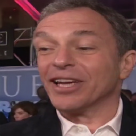} &
		\includegraphics{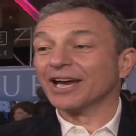} &
		\includegraphics{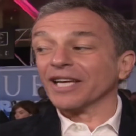}\\
		\includegraphics{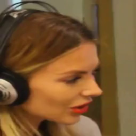} &
		\includegraphics{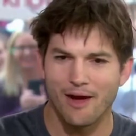}\hspace{1pt} &
		\includegraphics{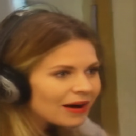} &
		\includegraphics{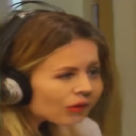} &
		\includegraphics{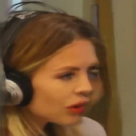} &
		\includegraphics{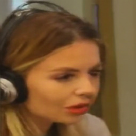} &
		\includegraphics{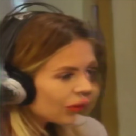} \\
		\includegraphics{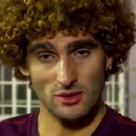} &
		\includegraphics{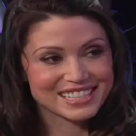}\hspace{1pt} &
		\includegraphics{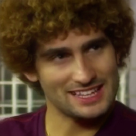} &
		\includegraphics{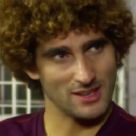} &
		\includegraphics{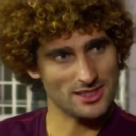} &
		\includegraphics{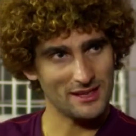} &
		\includegraphics{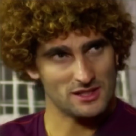}\\
		\includegraphics{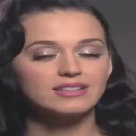} &
		\includegraphics{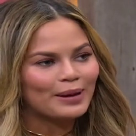}\hspace{1pt} &
		\includegraphics{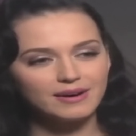} &
		\includegraphics{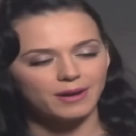} &
		\includegraphics{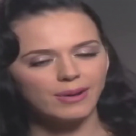} &
		\includegraphics{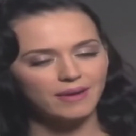} &
		\includegraphics{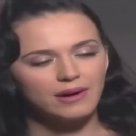}\\
		\includegraphics{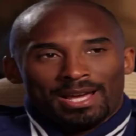} &
		\includegraphics{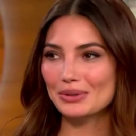}\hspace{1pt} &
		\includegraphics{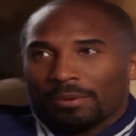} &
		\includegraphics{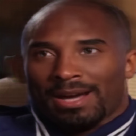} &
		\includegraphics{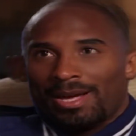} &
		\includegraphics{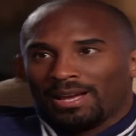} &
		\includegraphics{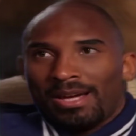}\\
		\includegraphics{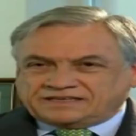} &
		\includegraphics{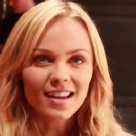}\hspace{1pt} &
		\includegraphics{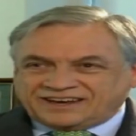} &
		\includegraphics{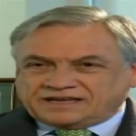} &
		\includegraphics{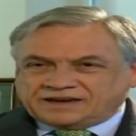} &
		\includegraphics{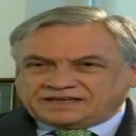} &
		\includegraphics{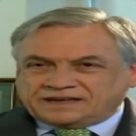}\\
		\includegraphics{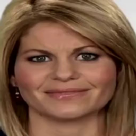} &
		\includegraphics{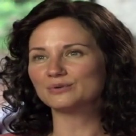}\hspace{1pt} &
		\includegraphics{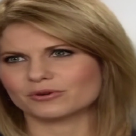} &
		\includegraphics{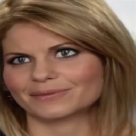} &
		\includegraphics{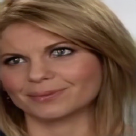} &
		\includegraphics{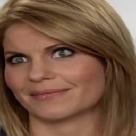} &
		\includegraphics{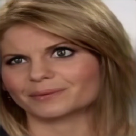}\\
		\includegraphics{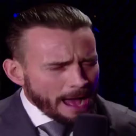} &
		\includegraphics{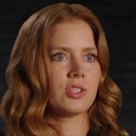}\hspace{1pt} &
		\includegraphics{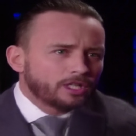} &
		\includegraphics{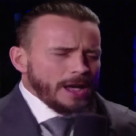} &
		\includegraphics{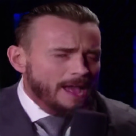} &
		\includegraphics{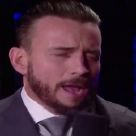} &
		\includegraphics{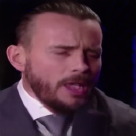}\\
	};
	
	\node[lu] at (m0.north) {\raisebox{1mm}[0mm][0mm]{\small Source}};
	\node[lu] at (m1.north) {\raisebox{1mm}[0mm][0mm]{\small Driving}};
	\node[lu] at (m2.north) {\raisebox{1mm}[0mm][0mm]{\small Ours}};
	\node[lu] at (m3.north) {\small TSMM~\cite{spline}};
	\node[lu] at (m4.north) {\small DaGAN~\cite{dagan}};
	\node[lu] at (m5.north) {\small OSFS \cite{one-shot}};
	\node[lu] at (m6.north) {\small FOMM \cite{fom}};

	\end{tikzpicture}\vspace{-1ex}
	\caption{
	Cross-reenactment generalization to driving videos and source images both from the VoxCeleb2 test set~\cite{chung18b} with relative motion transfer.
	}
	\label{fig:vox2}%
\end{figure*}

\newlength{\cidkpheight}\setlength{\cidkpheight}{2.06cm}
\newcommand{\imgcidkp}[1]{\includegraphics[height=\cidkpheight,clip,trim=0 0 0 0]{#1}}

\newcommand{\regimg}[1]{\includegraphics[height=2.16cm]{#1}}

\end{document}